%% file: www_2021.tex
\documentclass[sigconf]{acmart}

\usepackage[linesnumbered, ruled]{algorithm2e}
\usepackage[utf8]{inputenc}
\usepackage[english]{babel}
\usepackage{graphicx}  
\usepackage{subcaption}  
\usepackage{amsthm}
\usepackage{appendix}
\usepackage{mathtools}
\theoremstyle{definition}
\newtheorem{definition}{Definition}
\newtheorem{theorem}{Theorem}
\newtheorem{corollary}{Corollary}
\newtheorem{lemma}{Lemma}

\newtheorem{proposition}{Proposition}
\theoremstyle{remark}
\newtheorem*{remark}{Remark}

\usepackage{enumitem}
\newenvironment{hproof}{%
  \renewcommand{\proofname}{Proof Sketch}\proof}{\endproof}

\def \bx {\mathbf{x}}
\def \xij {\mathbf{x}_{ij}}
\def \xijt {{\mathbf{x}_{ij}^t}}
\def \xjit {{\mathbf{x}_{ji}^t}}

\def \xmns {{\mathbf{x}_{mn}^s}}
\def \xmnt {{\mathbf{x}_{mn}^t}}
\def \ymns {{\mathbf{y}_{mn}^s}}

\def \yijt {{\mathbf{y}_{ij}^t}}
\def \tp  {t^\prime}
\def \sub {\scriptscriptstyle}
\def \btheta {\boldsymbol{\theta}}
\def \bTheta {\boldsymbol{\Theta}}
\def \bR {\mathbb{R}}
\def \bM {\mathbf{M}}
\def \bZ {\mathbf{Z}}
\def \bU {\mathbf{U}}
\def \bG {\mathbf{G}}
\def \bSigma {\mathbf{\Sigma}}

\DeclareMathOperator*{\argmin}{argmin}

\AtBeginDocument{%
  \providecommand\BibTeX{{%
    \normalfont B\kern-0.5em{\scshape i\kern-0.25em b}\kern-0.8em\TeX}}}

\setcopyright{acmcopyright}
\copyrightyear{2021}
\acmYear{2021} 
\acmConference[WWW '21]{Proceedings of the Web Conference 2021}{April 19--23, 2021}{Ljubljana, Slovenia}
\acmBooktitle{Proceedings of the Web Conference 2021 (WWW '21), April 19--23, 2021, Ljubljana, Slovenia}
\acmPrice{}
\acmDOI{10.1145/3442381.3449972}
\acmISBN{978-1-4503-8312-7/21/04}

\begin{document}
\title{PairRank: Online Pairwise Learning to Rank by Divide-and-Conquer}
\newcommand{\model}{{PairRank}}

\author{Yiling Jia}
\affiliation{%
  \institution{University of Virginia}
  \city{Charlottesville}
  \state{VA}
  \country{USA}
}
\email{yj9xs@virginia.edu}

\author{Huazheng Wang}
\affiliation{%
  \institution{University of Virginia}
  \city{Charlottesville}
  \state{VA}
    \country{USA}
}
\email{hw7ww@virginia.edu}

\author{Stephen Guo}
\affiliation{%
  \institution{Walmart Labs}
  \city{Sunnyvale}
  \state{CA}
  \country{USA}
}
\email{sguo@walmartlabs.com}

\author{Hongning Wang}
\affiliation{%
  \institution{University of Virginia}
  \city{Charlottesville}
  \state{VA}
  \country{USA}
}
\email{hw5x@virginia.edu}


\begin{abstract}
Online Learning to Rank (OL2R) eliminates the need of explicit relevance annotation by directly optimizing the rankers from their interactions with users. However, the required exploration drives it away from successful practices in offline learning to rank, which limits OL2R's empirical performance and practical applicability.
In this work, we propose to estimate a pairwise learning to rank model online. In each round, candidate documents are partitioned and ranked according to the model's confidence on the estimated pairwise rank order, and exploration is only performed on the uncertain pairs of documents, i.e., \emph{divide-and-conquer}.  
Regret directly defined on the number of mis-ordered pairs is proven, which connects the online solution's theoretical convergence with its expected ranking performance. Comparisons against an extensive list of OL2R baselines on two public learning to rank benchmark datasets demonstrate the effectiveness of the proposed solution.
\end{abstract}



\maketitle

\input{introduction}
\input{related}
\section{Method}
\input{problem}
\input{algorithm}
\input{regret}

\input{experiment}

\section{Conclusion}
Existing OL2R solutions suffer from slow convergence and sub-optimal performance due to inefficient exploration and limited optimization strategies. Motivated by the success of offline models, we propose to estimate a pairwise learning to rank model on the fly, named as \model{}. 
Based on the model's pairwise order estimation confidence, exploration is performed only on the pairs where the ranker is still uncertain, i.e., \emph{divide-and-conquer}. 
We prove a sub-linear upper regret bound defined on the number of mis-ordered pairs, which directly links \model{}'s convergence with classical ranking evaluations.  
Our empirical experiments support our regret analysis and demonstrate significant improvement of \model{} over several state-of-the-art OL2R baselines. 

Our effort sheds light on moving more powerful offline learning to rank solutions online. Currently, our work is based on a single layer RankNet for analysis purposes. Following recent efforts of convergence analysis in deep learning \cite{zhou2019neural}, it is possible to extend \model{} with deep ranking models and directly optimize rank-based metrics (such as NDCG). Furthermore, most OL2R solutions focus on population-level ranker estimation; thanks to the improved learning efficiency by \model{}, it is possible for us to study individual-level ranking problems, e.g., personalized OL2R.

\section{Acknowledgements}
We want to thank the reviewers for their insightful comments. This work is based upon work supported by National Science Foundation under grant IIS-1553568 and IIS-1618948, and Google Faculty Research Award.


\bibliographystyle{ACM-Reference-Format}
\bibliography{reference}

\appendix

\input{supplementary}

\end{document}

%% file: introduction.tex
\section{Introduction}

Online learning to rank (OL2R) empowers modern retrieval systems to optimize their performance directly from users' implicit feedback \cite{wang2019variance,wang2018efficient,yue2009interactively,hofmann2013balancing,schuth2016multileave,zoghi2017online,lattimore2018toprank,oosterhuis2018differentiable,li2018online}. The essence of OL2R solutions is to infer the quality of individual documents under a given query \cite{radlinski2008learning,kveton2015cascading,zoghi2017online,lattimore2018toprank,li2018online}
or a parameterized ranking function \cite{wang2019variance,wang2018efficient,yue2009interactively,oosterhuis2018differentiable}
via sequential interactions with users, i.e., trial and error. It eliminates classical offline learning to rank solutions' strong dependency on explicit relevance annotations and makes supervised learning of ranking models possible when collecting explicit annotations from experts is economically infeasible or even impossible (e.g., private collection search). 

Although influential and theoretically sound, the current OL2R solutions are not compatible with the successful practices in offline learning to rank, which directly optimize rankers by minimizing loss defined by rank-based metrics, such as Average Relevance Position (ARP) \cite{joachims2017ips} or Normalized Discounted Cumulative Gain (NDCG) \cite{burges2010ranknet}. As a result, the performance of existing OL2R solutions is still behind that of offline solutions, which directly restricts OL2R's real-world applicability.

The key barrier separating the practices in online and offline learning to rank is the need of exploration. Since users' feedback is implicit and known to be biased and noisy \cite{joachims2005accurately,agichtein2006improving,joachims2007evaluating}, more clicks on a top-ranked document do not necessarily indicate greater relevance. Effective exploration in the problem space is thus vital for the online model update. 
Current OL2R solutions explore either in the action space (e.g., presenting currently underestimated results at higher positions of a ranked list) \cite{radlinski2008learning,kveton2015cascading,zoghi2017online,lattimore2018toprank}, or in the model space (e.g., presenting ranked results from different rankers)  \cite{yue2009interactively,schuth2014multileaved}. 
However, due to the combinatorial nature of ranking, the action space is too large to be efficiently explored (e.g., all permutations of returned documents). This forces such OL2R solutions to take a \emph{pointwise} approach to estimate the utility of each query-document pair separately, which has proven to be inferior to the pairwise or listwise approaches in offline learning to rank studies \cite{chapelle2011yahoo}. While for model space exploration,
though an interleaved test makes it possible to compare different rankers with respect to a hidden utility function in an unbiased fashion, it is hard to link this comparison to the optimization of any rank-based metrics. Moreover, due to the required uniform sampling in the model space, this type of OL2R solutions suffers from high variance and high regret during online result serving and model update \cite{wang2019variance}. 

In this work, we aim to bridge the gap by directly training a \emph{pairwise} learning to rank model online. We target pairwise ranking models for three major reasons. First, a pairwise ranker reduces the exponentially sized action space to quadratic, by deriving the full ranking order from the pairwise comparisons between documents. Second, existing studies in search log analysis demonstrate relative preferences derived from clicks are more accurate and reliable than absolute judgments \cite{joachims2005accurately,joachims2007evaluating}. Third, pairwise learning to rank models have competitive empirical performance and have been widely used in practical systems \cite{chapelle2011yahoo,joachims2002optimizing,burges2010ranknet}. 
To gain theoretical insight into the proposed solution, we work with a single layer RankNet model with a sigmoid activation function \cite{burges2010ranknet}, which makes analytical convergence analysis possible. In particular, we explore in the pairwise ranking space of all candidate documents via \emph{divide-and-conquer}:

we partition documents under a given query in each round of result serving, where the document ranking across different parts of the partition is certain (e.g., all documents in one part should be ranked higher than those in another part), but the ranking among documents within each part is still uncertain.
The ranked list is thus generated by a topological sort across parts of a partition and randomly shuffling within each part. We name our solution \model{}.
We rigorously prove that the exploration space shrinks exponentially fast in \model{} as the ranker estimation converges, such that the cumulative regret defined on the number of mis-ordered pairs has a sublinear upper bound. As most existing ranking metrics can be reduced to different kinds of pairwise comparisons among candidate documents \cite{Wang2018Lambdaloss}, e.g., Average Relevance Position is counted over a relevant document against all other documents, \model{} can directly optimize those ranking metrics based on users' implicit feedback on the fly. Our extensive empirical evaluations also demonstrate the strong advantage of \model{} against a rich set of state-of-the-art OL2R solutions over a collection of OL2R benchmark datasets on standard retrieval metrics.  

%% file: related.tex
\section{Related Work}

We broadly categorize existing OL2R solutions into two families.
The first type learns the best ranked list for each individual query separately by modeling users' click and examination behaviors with multi-armed bandit algorithms \cite{radlinski2008learning,kveton2015cascading,zoghi2017online,lattimore2018toprank}. 
Typically, solutions in this category depend on specific click models to decompose the estimation on each query-document pair; as a result, exploration is performed on the ranking of individual documents. For example, by assuming users examine documents from top to bottom until reaching the first relevant document,  cascading bandit models perform exploration by ranking the documents based on the upper confidence bound of their estimated relevance \cite{kveton2015cascading, kveton2015combinatorial, li2016contextual}. 
Other types of click models have also been explored (such as the dependent click model) \cite{katariya2016dcm,zoghi2017online, lattimore2018toprank, li2018online, kveton2018bubblerank}. 
However, as the relevance is estimated for each query-document pair, such algorithms can hardly generalize to unseen queries or documents. Moreover, pointwise relevance estimation is proved to be ineffective for rank estimation in established offline learning to rank studies \cite{chapelle2011yahoo,burges2010ranknet}.

The second type of OL2R solutions leverage ranking features for relevance estimation and explores for the best ranker in the entire model space \cite{yue2009interactively,li2018online,oosterhuis2018differentiable}. The most representative work is Dueling Bandit Gradient Descent (DBGD) \cite{yue2009interactively,schuth2014multileaved}, which proposes an exploratory direction in each round of interaction and uses an interleaved test \cite{chapelle2012large} to validate the exploration for model update. To ensure an unbiased gradient estimate, DBGD uniformly explores the entire model space, which 
costs high variance and high regret during online ranking and model update. 
Subsequent methods improved upon DBGD by developing more efficient sampling strategies, such as multiple interleaving and projected gradient, to reduce variance \cite{hofmann2012estimating,zhao2016constructing,oosterhuis2017balancing, wang2018efficient, wang2019variance}. However, as exploration is performed in the model space, click feedback is used to infer which ranker is preferred under a hypothetical utility function. It is difficult to reason how the update in DBGD is related to the optimization of any rank-based metric. Hence, though generalizable, this type of OL2R solutions' empirical performance is still worse than classical offline solutions.



The clear divide between the practices in online and offline learning to rank is quite remarkable, which has motivated some recent efforts to bridge the gap. \citet{hofmann2013balancing} adopt $\epsilon$-greedy to estimate a stochastic RankSVM~\cite{joachims2002optimizing, herbrich1999support} model on the fly. Though RankSVM is effective for pairwise learning to rank, the totally random exploration by $\epsilon$-greedy is independent from the learned ranker. It keeps distorting the ranked results, even when the ranker has identified some high-quality results. \citet{oosterhuis2018differentiable} perform exploration by sampling the next ranked document from a Plackett-Luce model and estimate gradients of this ranking model from the inferred pairwise result preferences. Although exploration is linked to the ranker's estimation, the convergence of this solution is still unknown. 


%% file: problem.tex
In this section, we present our solution, which trains a pairwise learning to rank model online. The key idea is to partition the pairwise document ranking space and only explore the pairs where the ranker is currently uncertain, i.e., divide-and-conquer. We name our solution \model{}. We rigorously prove the regret of \model{} defined on the cumulative number of mis-ordered pairs over the course of online model update. 

\subsection{Problem Formulation}
\label{sec:problem}


In OL2R, a ranker interacts with users for $T$ rounds. At each round $t=1,2,...,T$, the ranker receives a query $q_t$ and its associated candidate documents, 
which are represented as a set of $d$-dimensional query-document feature vectors $\mathcal{X}_t = \{\bx_1^t, \bx_2^t, \dots \bx_{L_t}^t\}$ with $\bx^t_i \in \bR^d$ and $\Vert\bx_i^t\Vert \leq u$. The ranker determines the ranking of the candidate documents $\tau_t = \big(\tau_t(1), \tau_t(2), \dots, \tau_t(L_t)\big) \in \Pi([L_t])$, based on its knowledge so far, where $\Pi([L_t])$ represents the set of all permutations of $L_t$ documents and $\tau_t(i)$ is the rank position of document $i$ under query $q_t$. Once the ranked list is returned to the user, the user examines the results and provides his/her click feedback $C_t = \{c_1^t, c_2^t, ..., c_{L_t}^t\}$, where $c_i^t = 1$ if the user clicks on document $i$ at round $t$; otherwise $c_i^t = 0$. Based on this feedback, the ranker updates itself and proceeds to the next query.

$C_t$ is known to be biased and noisy \cite{joachims2005accurately,agichtein2006improving,joachims2007evaluating}. Existing studies find that users tend to click more on higher-ranked documents, as they stop examination early in the list. This is known as position bias. And users can only interact with the documents shown to them, known as the presentation bias. As a result, the ranker cannot simply treat non-clicked documents as irrelevant. 
Such implicit feedback imposes several key challenges for online learning to rank: how to deal with the implicit feedback, and in the meanwhile, how to effectively explore the unknowns for the model update. 
Following the practice in \cite{joachims2005accurately}, we treat clicks as relative preference feedback and assume that clicked documents are preferred over the \emph{examined but unclicked} ones. In addition, we consider every document that precedes a clicked document and the first subsequent unclicked document as examined. This approach has been widely adopted and proven to be effective in learning to rank \cite{wang2019variance,agichtein2006improving,oosterhuis2018differentiable}. Accordingly, we use $o_t$ to represent the index of the last examined position in the ranked list $\tau_t$ at round $t$.

Exploration is the key component that differentiates OL2R to offline L2R, where OL2R needs to serve while learning from its presented rankings. 
The most straightforward exploration is to provide a random list of candidate documents. However, such random exploration is less appreciated for OL2R as it hurts user experience, even though it may be beneficial for model training.
Therefore, regret becomes an important metric for evaluating OL2R.
Though various types of regret have been defined and analyzed in existing OL2R studies, few of them link to any rank-based metric, which is the key in ranker evaluation. For example, for OL2R solutions that explore in the document space, regret is typically defined on the number of clicks received on the presented ranking versus that known only in hindsight \cite{li2016contextual,kveton2015cascading,zoghi2017online,lattimore2018toprank}. For solutions that explore in the model space, such as DBGD, regret is defined as the number of rounds where the chosen ranker is preferred over the optimal ranker \cite{yue2009interactively,wang2019variance}. It is difficult to reason how such measures indicate an OL2R solution's ranking performance against a desired retrieval metric, such as ARP and NDCG.
To bridge this gap, we define regret by the number of mis-ordered pairs from the presented ranking to the ideal one, i.e., the Kendall tau rank distance,
\begin{equation*}
    R_T = \mathbb{E}\big[\sum\nolimits_{t=1}^T r_t\big] = \mathbb{E} \big[\sum\nolimits_{t=1}^T K(\tau_t, \tau_t^*)\big]
\end{equation*}
where $K(\tau_t, \tau_t^*)=\Big|\big\{(i,j):i<j,\big(\tau_{t}(i)<\tau _{t}(j)\wedge \tau^*_{t}(i)>\tau^*_{t}(j)\big)\vee \big(\tau _{t}(i)>\tau _{t}(j)\wedge \tau^*_{t}(i)<\tau^*_{t}(j)\big)\big\}\Big|$. As shown in \cite{Wang2018Lambdaloss}, most ranking metrics employed in real-world retrieval systems, such as ARP and NDCG, can be decomposed into pairwise comparisons; hence, our defined regret directly connects an OL2R algorithm's online performance with classical ranking evaluations.

%% file: algorithm.tex
\subsection{Online Pairwise Learning to Rank}
\label{sec_model}

\begin{figure}[t]
  \vspace{-2mm}
  \centering
  \includegraphics[width=\linewidth]{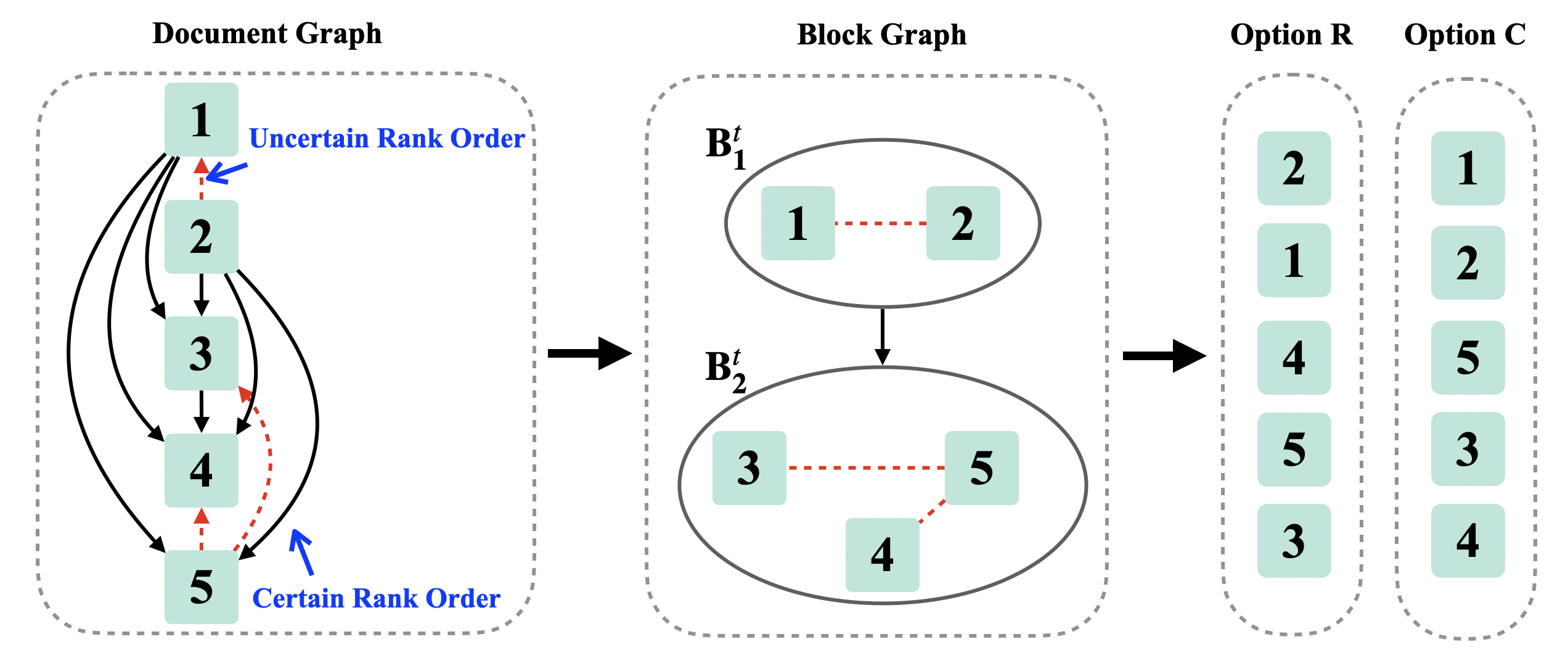}
  \Description[An illustration of \model{} model]{From document graph to block graph, and then the generated ranked list.}
  \caption{PairRank: pairwise explore and exploit by divide-and-conquer. Assume that the optimal ranking order among the 5 documents is $1 \succ 2 \succ 3 \succ 4 \succ 5$. At the current round $t$, the ranker is confident about its preference estimation between all the pairs expect $(1, 2), (3, 5), (4 ,5)$. In this example, the instantaneous regret of the first proposed ranked list is 3 and 2 in the second proposal.}
  \label{fig:model}
  \vspace{-2mm}
\end{figure}

The key in OL2R is to effectively explore the unknowns while providing high-quality results to the users, which is often referred to as the explore-exploit trade-off. In this work, we propose to directly train a pairwise model from its interactions with users and directly explore in the pairwise document ranking space via a divide-and-conquer strategy. The high-level idea of \model{} is illustrated in Figure \ref{fig:model}, and we explain its details in the following.

\subsubsection{Pairwise Learning to Rank}

We focus on the pairwise ranking models because of their advantageous effectiveness reported in prior offline LTR studies \cite{chapelle2011yahoo}, and their ability to tackle implicit feedback.
Specifically, we adopt a single layer RankNet model with a sigmoid activation function \cite{burges2010ranknet} as our pairwise ranker.
This choice is based on the promising empirical performance of RankNet and the feasibility of analyzing the resulting online solution's convergence. 

In a single layer RankNet, the probability that a document $i$ is more relevant than document $j$ under query $q$ is computed as $\mathbb{P}(i \succ j | q) = \sigma({\bx^\top_i}\btheta - {\bx^\top_j}\btheta)$, where $\btheta \in \bR^d$ and $\Vert\btheta\Vert \leq Q$ is the model parameter and $\sigma(x)= {1}/({1 + \exp(-x)})$.
To simplify our notations, we use $\xij$ to denote $\bx_i - \bx_j$ in our subsequent discussions.

With the knowledge of $\btheta$, due to the monotonicity and transitivity of the sigmoid function, the ranking of documents in $\mathcal{X}$ can be uniquely determined by $\{{\bx^\top_1}\btheta, {\bx^\top_2}\btheta, \dots, {\bx^\top_{L}}\btheta\}$. Therefore, the key of learning a RankNet model is to estimate its parameter $\btheta$. As RankNet specifies a distribution on pairwise comparisons, the objective function for $\btheta$ estimation can be readily derived as the cross-entropy loss between the predicted pairwise distribution on documents and those inferred from user feedback till round $t$:
\small
\begin{align}
\label{eqn:loss}
    \mathcal{L}_t = \sum_{s=1}^t\sum_{(m, n) \in \mathcal{G}_{s}} & - \ymns \log\big(\sigma({\xmns}^\top\btheta)\big)  \\
    & - (1 - \ymns)\log\big(1 - \sigma({\xmns}^\top\btheta)\big) + \frac{\lambda}{2} \Vert\btheta\Vert^2 \nonumber 
\end{align}
\normalsize
where $\lambda$ is the L2 regularization coefficient, $\mathcal{G}_{s}$ denotes the set of document pairs that received different click feedback at round $s$, i.e., $\mathcal{G}_{s} = \{(m, n): c_m^{s} \neq c_n^{s}, \forall \tau(m) < \tau(n) \leq o_{s}\}$, $\ymns$ indicates whether the document $m$ is preferred over document $n$ in the click feedback, i.e., $\ymns = \frac{1}{2}(c_m^t - c_n^t) + \frac{1}{2}$~\cite{burges2010ranknet}. Due to the log-convexity of the loss function defined in Eq~\eqref{eqn:loss}, the global optimal solution $\hat{\btheta}_t$ at round $t$ exists and can be efficiently obtained by gradient descent. 
 
Online learning of RankNet boils down to the construction of $\{\mathcal{G}_{s}\}^T_{s=1}$ over time. However, the conventional practice of using all the inferred pairwise preferences from click feedback \cite{joachims2002optimizing,agichtein2006improving} imposes a higher risk in an online setting. In the presence of click noise (e.g., a user mistakenly clicks on an irrelevant document), pairing documents would cause a quadratically increasing number of noisy training instances, which impose strong negative impact on the quality of the learned ranker \cite{carvalho2008suppressing}. As the updated ranker is immediately executed, cascading of ranker deterioration is possible. 
To alleviate this deficiency, we propose to only use independent pairs inferred from the feedback, e.g., $\mathcal{G}_{s}^{ind} = \{(m, n): c_m^{s} \neq c_n^{s}, \forall (\tau_s(m), \tau_s(n)) \in D\}$, where $D$ represents the set of disjointed position pairs, for example, $D = \{(1, 2), (3, 4), ... (L-1, L)\}$. In other words, we will only use a subset of non-overlapping pairwise comparisons for our online ranker update. 

\subsubsection{Uncertainty Qualification}
\begin{figure}[t]
  \centering
  \includegraphics[width=0.8\linewidth]{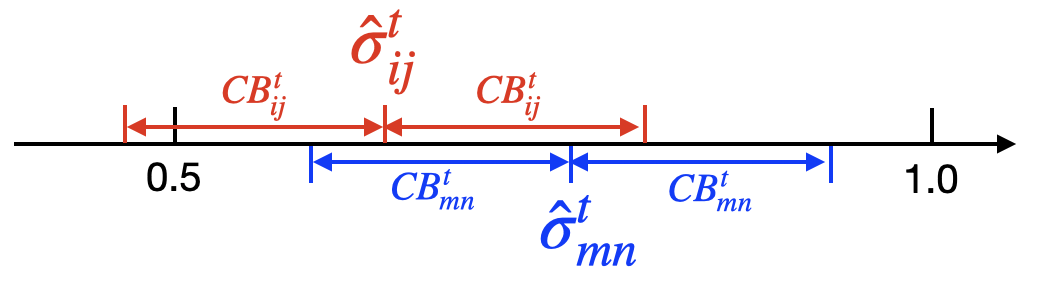}
  \Description[Comparison between certain and uncertain rank order]{How estimation uncertainty determines the certain and uncertain rank order.}
  \vspace{-2mm}
  \caption{Illustration of certain and uncertain rank orders.}
  \label{fig:certain_rank}
  \vspace{-2mm}
\end{figure}

As discussed in Section \ref{sec:problem}, $\hat{\btheta}_t$ is obtained based on the acquired feedback from what has been presented to the user, which is subject to various types of biases and noises \cite{joachims2005accurately,agichtein2006improving,joachims2007evaluating}. Hence, $\hat{\btheta}_t$ only reflects what the ranker knows so far; and it is vital to effectively explore the unknowns to complete its knowledge. In \model{}, we propose to explore in the pairwise document ranking space spanned by $\mathcal{X}_t$ under $q_t$, with respect to the current ranker's uncertainty about the pairwise comparisons.

Accurate quantification of the current ranker's uncertainty on pairwise preference estimation is the key to such exploration. The model estimation uncertainty is caused by the existence of click noise, i.e., $\Vert\hat{\btheta}_t - \btheta^*\Vert \neq 0$, where $\btheta^*$ is the underlying ground-truth model parameter. And this model estimation uncertainty directly leads to the uncertainty in the ranker's pairwise preference estimation. 
To quantify the source of uncertainty, we follow conventional click models to assume the clicks are independent of each other given the true relevance of documents, so as their noise \cite{joachims2005accurately,guo2009click,guo2009efficient}. As a result, the pairwise noise becomes the sum of noise from the two associated clicks. Because we only use the independent document pairs $\mathcal{G}^{ind}$, the pairwise noise is thus independent of each other in \model{} and the history of result serving, which directly leads to the following proposition.

\begin{proposition}
For any $t \geq 1$, $\forall (i, j) \in \mathcal{G}_t^{ind}$, define pairwise noise $\epsilon_{ij}^t = \yijt - \sigma(\xijt^\top\theta^*)$. For all $t \geq 1$, $\epsilon_{ij}^t$ is a sub-Gaussian random variable with $\mathbb{E}[\epsilon_{ij}^t| \mathcal{G}_{t-1}^{ind}, \{\epsilon^{t-1}\}, \dots, \mathcal{G}_{1}^{ind}, \{\epsilon^{1}\}] = 0$, where $\{\epsilon^t\} = \{\epsilon_{ij}^t, (i, j)\in\mathcal{G}_t^{ind}\}$
\end{proposition}

According to the property of sub-Gaussian random variables, such assumption can be easily satisfied in practice as long as the pointwise click noise follows a sub-Gaussian distribution. For example, the pointwise noise can be modeled as a binary random variable related to the document's true relevance under the given query, which follows $\frac{1}{2}$-sub-Gaussian distribution.

Therefore, based on the solution of Eq~(\ref{eqn:loss}), the uncertainty of the estimated pairwise preference $\sigma({\bx_{ij}^t}^\top \hat{\btheta}_t)$ by RankNet at round $t$ can be analytically bounded with a high probability, as shown in the following lemma.

\begin{lemma} (Confidence Interval of Pairwise Preference Estimation). At rount $t < T$, for any pair of documents $\bx_i^t$ and $\bx_j^t$ under query $q_t$, with probability at least $1 - \delta_1$, we have,
\begin{equation*}
    |\sigma({\xijt}^\top \hat{\btheta}_t) - \sigma({\xijt}^\top \btheta^*) | \leq \alpha_t\Vert\xijt\Vert_{\mathbf{M}_t^{-1}},
\end{equation*}
where $\alpha_t = ({2k_{\mu}}/{c_{\mu}}) \Big(\sqrt{R^2\log{({\det(\mathbf{M}_t)}/({\delta_1^2 \det(\lambda \mathbf{I})})})} + \sqrt{\lambda} Q\Big)$, $\mathbf{M}_t = \lambda \mathbf{I} + \sum_{s =1}^{t-1}\sum_{(m, n) \in \mathcal{G}^{ind}_{s}}\xmns\xmns^\top$, $k_{\mu}$ is the Lipschitz constant of the sigmoid link function $\sigma$, $c_{\mu} = \inf_{\btheta \in \bTheta} \dot{\sigma}(\bx^\top\btheta)$, with $\dot{\sigma}$ as the first derivative of $\sigma$, and $R$ is the sub-gaussian parameter for noise $\epsilon$.
\label{lemma:cb}
\end{lemma}

The detailed proof of Lemma~\ref{lemma:cb} can be found in the appendix. This lemma provides a tight high probability bound of the pairwise preference estimation uncertainty under a RankNet specified by $\hat\btheta_t$, which enables us to perform efficient pairwise exploration for model update. To better illustrate our exploration strategy based on the pairwise estimation uncertainty, we introduce the following definition on document pairs.

\begin{definition} (Certain Rank Order)
At any round $t < T$, the ranking order between document $i$ and $j$, denoted as $(i, j)$, is considered in a certain rank order if and only if $\sigma({\bx_{ij}^t}^\top\hat{\btheta}_t) - \alpha_{t}\Vert\bx_{ij}^t\Vert_{\mathbf{M}_t^{-1}} > \frac{1}{2}$.
\end{definition}

Intuitively, based on Lemma~\ref{lemma:cb}, if $(i, j)$ is in a certain rank order, with a high probability that the estimated preference (order) between document $i$ and $j$ is consistent with the ground-truth. 
For example, as shown in Figure~\ref{fig:certain_rank}, $\hat{\sigma}^t_{ij}$ and $\hat{\sigma}_{mn}^t$ represent the estimated pairwise preference on document pair $(i,j)$ and $(m,n)$ based on $\hat{\btheta}_t$, while $CB_{ij}^t$ and $CB_{mn}^t$ represent the corresponding confidence bound defined in Lemma~\ref{lemma:cb}, i.e., $CB_{ij}^t = \alpha_t\Vert\xijt\Vert_{\mathbf{M}_t^{-1}}$ and $CB_{mn}^t = \alpha_t\Vert\xmnt\Vert_{\mathbf{M}_t^{-1}}$. According to Lemma~\ref{lemma:cb},  we know that the ground-truth pairwise preferences, $\sigma_{ij}^*$ and $\sigma_{mn}^*$, lie within the corresponding confidence intervals with a probability at least $1 - \delta_1$, i.e., $\sigma_{ij}^* \in [\hat{\sigma}_{ij}^t - CB_{ij}^t, \hat{\sigma}_{ij}^t + CB_{ij}^t]$. 
In Figure~\ref{fig:certain_rank}, for pair $(m, n)$, the lower bound of its pairwise estimation, $\hat{\sigma}_{mn}^t - CB_{mn}^t$, is greater than $\frac{1}{2}$. This indicates that with a high probability $1 - \delta_1$, the estimated preference between document $m$ and $n$ is consistent with the ground-truth model $\btheta^*$; and thus there is no need to explore this pair. In contrast, with $\hat{\sigma}_{ij}^t - CB_{ij}^t < \frac{1}{2}$, the estimated order $(i \succ j)$ is still with uncertainty as the ground-truth model may present an opposite order; hence, exploration on this pair is necessary. 

We use $\mathcal{E}_{c}^t$ to represent the set of all certain rank orders at round $t$. Accordingly, the set of uncertain rank orders at round $t$ is defined as: $\mathcal{E}_{u}^t = \{(i, j) \in [L_t]^2: (i, j) \notin \mathcal{E}_c^t \wedge (j, i) \notin \mathcal{E}_c^t\}$. 

\subsubsection{Explore the Unknowns via Divide-and-Conquer}

\begin{algorithm}[t]
\caption{PairRank}
\label{algo:algo1}
    \textbf{Input:} $\lambda$, $\delta_1$, $\delta_2$
    Initialize $\mathbf{M}_0 = \lambda \mathbf{I}, \hat{\btheta}_1 = 0$
    \For{$t=1$ \KwTo $T$}{
        \text{Receive query $q_t$ and its corresponding candidate documents set} $\mathcal{X}_t = \{\bx_1^t, \bx_2^t, ..., \bx_{L_t}^t\}$. \\
        
        $\mathcal{E}_c^t = \{(i, j) \in [L_t]^2: \sigma({\bx_{ij}^t}^\top\mathbf{\hat{\btheta}}_{t-1}) - \alpha_{t-1}\Vert\bx_{ij}^t\Vert_{\mathbf{M}_{t-1}} > 1/2\}$ \\
        $\mathcal{E}_u^t = \{(i, j) \in [L_t]^2: (i, j) \notin \mathcal{E}_c^t \wedge (j, i) \notin \mathcal{E}_c^t\}$ \\
        \text{Construct ordered block list} $\mathcal{B}_t = \{\mathcal{B}_1^t, \mathcal{B}_2^t, ... \mathcal{B}_{d_t}^t\}$ \\
        \text{Generate ranked list} $\tau_t = \{\pi(\mathcal{B}_1^t), \pi(\mathcal{B}_2^t), ... , \pi(\mathcal{B}_{d_t}^t) \}$  \\
        \text{Observe click feedback} ${C}_t$, and corresponding $o_t$. \\
        $\mathcal{G}_t^{ind} = \{(i, j) \in [o_t]^2: c_i^t \ne c_j^t \wedge (\tau_t(i), \tau_t(j)) \in D\}$. \\
        \text{Estimate $\hat{\theta}_t$ as the solution of Eq \eqref{eqn:loss}}. \\
        $\mathbf{M}_t = \mathbf{M}_{t-1} + \sum_{(i, j) \in \mathcal{G}_t^{ind}} \mathbf{x}_{ij}^t {\mathbf{x}_{ij}^t}^\top$.
    }
\end{algorithm}

With the aforementioned pairwise estimation uncertainty and the corresponding sets of certain and uncertain rank orders, i.e., $\mathcal{E}_c^t$ and $\mathcal{E}_u^t$, we can effectively explore the unknowns.
Intuitively, we only need to randomize the ranking of documents among which the model is still uncertain about their ranking orders, i.e., the uncertain rank orders, and therefore obtain feedback to further update the model (and reduce uncertainty). 
For example, in the document graph shown in Figure \ref{fig:model}, the solid lines represent certain rank orders, while the dash lines represent the uncertain rank orders. When generating the ranked list, we should randomly swap the order between document 1 and 2 (i.e., to explore) while preserving the order between document 1 and documents 3, 4, 5 (i.e., to exploit).

This naturally leads to a divide-and-conquer exploration strategy in the space of pairwise document comparisons. Specifically, we partition $\mathcal{X}_t$ into different parts (referred to as \textit{blocks} hereafter) based on $\mathcal{E}_c^t$ and $\mathcal{E}_{u}^t$ such that the ranking orders between any documents belonging to different blocks are certain, shown in the block graph in Figure~\ref{fig:model}. The ranked list can thus be generated by topological sort across blocks and random shuffling within each block. As the exploration is confined to the pairwise ranking space, it effectively reduces the exponentially sized action space to quadratic. 

Algorithm~\ref{algo:algo1} shows the detailed steps of \model{}. At round t, we first construct $\mathcal{E}_c^t$ and $\mathcal{E}_u^t$ according to the current mode $\hat{\btheta}_{t-1}$. Then, we create blocks of $\mathcal{X}_t$ according to the definition below.

\begin{definition} (Block)
At any round $t < T$, the block $\mathcal{B}$ is a set of documents that satisfy: 
\begin{enumerate}[nolistsep]
    \item $\forall i \in \mathcal{B}_d^t, (i, j) \in \mathcal{E}_c^t \text{ for any } j\in[L_t]\setminus \mathcal{B}_d^t.$ 
    \item $\nexists k \in [L_t] \setminus \mathcal{B}_d^t,  \text{for } i, j \in \mathcal{B}_d^t, (i, k) \in \mathcal{E}_t^c \wedge (k, j) \in \mathcal{E}_c^t$
\end{enumerate}
\end{definition}

Intuitively, each block is a subset of documents linked to each other by the \textit{uncertain rank order}. It can be viewed as a connected component in an undirected graph with documents as vertices and \textit{uncertain rank order} as edges under a given query. The connected components (blocks) can be found by linearly scanning through the vertices, based on breadth-first search or depth-first search if a vertex is not visited before. When the algorithm stops, each vertex (document) will be assigned to a connected component (block). Once the blocks are constructed, the order of the blocks can be obtained by topological sort (line 6). Let $\mathcal{B}_t = \{\mathcal{B}_1^t, \mathcal{B}_2^t, ... \mathcal{B}_{d_t}^t\}$ be the ordered block list for $\mathcal{X}_t$ at round $t$, the ranked list $\tau_t$ is generated as $\tau_t = \{\pi(\mathcal{B}_1^t), \pi(\mathcal{B}_1^t), ... , \pi(\mathcal{B}_{d_t}^t) \}$, where $\pi(\cdot)$ randomly permutes its input set as output.

To further improve exploration efficiency, we propose two options to generate the ranked list. As shown in Figure~\ref{fig:model}, the first ranked list is generated by randomly shuffling all documents within each block (referred to as random exploration), while in the second list, only the uncertain rank orders are shuffled, and the certain ones are preserved (referred to as conservative exploration). In our empirical studies, we observe such conservative exploration gives better improvement than random exploration, which further confirms the importance of efficient exploration in OL2R.

%% file: regret.tex
\subsection{Regret Analysis}

We theoretically analyze the regret of \model{}, which is defined by the cumulative number of mis-ordered pairs in its proposed ranked list till round $T$. 
The key in analyzing this regret is to quantify how fast the model achieves certainty about its pairwise preference estimation in candidate documents. First, we define $E_t$ as the success event at round $t$:
\begin{equation*}
    E_t = \big\{ \forall (i, j) \in [L_t]^2, |\sigma({\mathbf{x}_{ij}^t}^\top\hat{\btheta}_t) - \sigma({\mathbf{x}_{ij}^t}^\top\btheta^*) | \leq \alpha_t\Vert\mathbf{x}_{ij}^t\Vert_{\bM_t^{-1}}\big\}.
\end{equation*}

Intuitively, $E_t$ is the event that the estimated $\hat{\theta}_t$ is ``close'' to the optimal model $\theta^*$ at round $t$. According to Lemma~\ref{lemma:cb}, it is easy to reach the following conclusion,

\begin{corollary}On the event $E_t$, it holds that $\sigma({\mathbf{x}_{ij}^t}^\top\btheta^*) > {1}/{2}$ if $(i, j) \in \mathcal{E}_c^t$.
\label{col}
\end{corollary} 

\model{} suffers regret as a result of misplacing a pair of documents, i.e., swapping a pair already in the correct order. Based on Corollary~\ref{col}, under event $E_t$, the certain rank order identified by \model{} is consistent with the ground-truth. 
As our partition design always places documents under a certain rank order into distinct blocks, under event $E_t$ the ranking order across blocks is consistent with the ground-truth. In other words, regret only occurs when ranking documents within each block.


To analyze the regret caused by random shuffling within each block, we need the following technical lemma derived from random matrix theory. We adapted it from Equation (5.23) of Theorem 5.39 from \cite{vershynin2010introduction}.

\begin{lemma}
\label{lemma:matrix}
Let $A \in \mathbb{R}^{n \times d}$ be a matrix whose rows $A_i$ are independent sub-Gaussian isotropic random vectors in $\mathbb{R}^d$ with parameter $\sigma$, namely $\mathbb{E}[\text{exp}(x^\top (A_i - \mathbb{E}[A_i])] \leq \text{exp}(\sigma^2\Vert x \Vert ^2 / 2)$ for any $x \in \mathbb{R}^d$. Then, there exist positive universal constants $C_1$ and $C_2$ such that, for every $t \geq 0$, the following holds with probability at least $1 - 2\text{exp}(-C_2t^2), \text{where } \epsilon = \sigma(C_1\sqrt{d/n} + {t}/{\sqrt{n}})$: $\Vert A^\top A/n - \mathbf{I}_d\Vert \leq \text{max}\{\epsilon, \epsilon^2\}$.
\end{lemma}

The detailed proof can be found in \cite{vershynin2010introduction}. We should note the condition in Lemma~\ref{lemma:matrix} is not hard to satisfy in OL2R: at every round, the ranker is serving a potentially distinct query; and even for the same query, different documents might be returned at different times. This gives the ranker a sufficient chance to collect informative observations for model estimation.  Based on Lemma \ref{lemma:matrix}, we have the following lemma, which provides a tight upper bound of the probability that \model{}'s estimation of the pairwise preference is an uncertain rank order.

\begin{lemma}
At round $t \geq t^\prime$, with $\delta_1 \in (0, \frac{1}{2})$, $\delta_2 \in (0, \frac{1}{2})$, $\beta \in (0, \frac{1}{2})$, and $C_1$, $C_2$ defined in Lemma~\ref{lemma:matrix}, under event $E_t$, the following holds with probability at least $1 - \delta_2$: $\forall (i, j) \in [L_t]^2$, $\mathbb{P}\big((i, j) \in \mathcal{E}_u^t\big) \leq \frac{8k_{\mu}^2\Vert\xijt\Vert_{\bM_t^{-1}}^2}{(1 - 2\beta) c_{\mu}^2\Delta_{\min}^2}\log{\frac{1}{\delta_1}}$, with 
$t^\prime = \Big(\frac{c_1\sqrt{d} + c_2\sqrt{\log(\frac{1}{\delta_2})} + abd\sqrt{\frac{o_{\text{max}}u^2}{d\lambda}}}{\lambda_{\text{min}}(\bSigma)}\Big)^2 + \frac{2ab\log({1}/{\delta_1^2}) + 8a\lambda Q^2 - \lambda}{\lambda_{\text{min}}(\bSigma)}$, 
where $\Delta_{\min} = \min\limits_{t\in T, (i, j) \in [L_t]^2}| \sigma({\xijt}^\top\btheta^*) - \frac{1}{2}|$ representing the smallest gap of pairwise difference between any pair of documents associated to the same query over time (across all queries), $a = {4k_{\mu}^2 u^2}/({\beta^2c_{\mu}^2\Delta_{\text{min}}^2})$, and $b = R^2 + 4\sqrt{\lambda}QR$, 
\label{lemma:uncertain}
\end{lemma}

\begin{hproof}
According to the definition of certain rank order, a pairwise estimation $\sigma(\xijt^\top\hat{\btheta})$ is certain if and only if $|\sigma(\xijt^\top\hat{\btheta}) - 1/2| \geq \alpha_t\Vert \xijt\Vert_{\mathbf{M}_{t}^{-1}}$. By the reverse triangle inequality, the probability can be upper bounded by $\mathbb{P}\big(\big| |\sigma(\xijt^\top\hat{\btheta}) - \sigma(\xijt^\top\btheta^*)| - |\sigma(\xijt^\top\btheta^*) - 1/2|\big| \geq \alpha_t\Vert \xijt\Vert_{\mathbf{M}_{t}^{-1}}\big)$, which can be further bounded by Theorem 1 in \cite{abbasi2011improved}. The key in this proof is to obtain a tighter bound on the uncertainty of \model{}'s parameter estimation compared to the bound determined by $\delta_1$ in Lemma~\ref{lemma:cb}, such that its confidence interval on a pair of documents' comparison at round $t$ will exclude the possibility of flipping their ranking order, i.e., the lower confidence bound of this pairwise estimation is above 0.5.
\end{hproof}

In each round of result serving, as the model $\hat{\btheta}_t$ would not change until next round, the expected number of uncertain rank orders, denoted as $N_t=|\mathcal{E}_{u}^t|$, can be estimated by the summation of the uncertain probabilities over all possible pairwise comparisons under the current query $q_t$, e.g., $\mathbb{E}[N_t] = \frac{1}{2} (\sum_{(i, j) \in [L_t]^2}  \mathbb{P}((i, j) \in \mathcal{E}_u^t)$.

Denote $p_{t}$ as the probability that the user examines all documents in $\tau_t$ at round $t$, and let $p^* = \min_{1\leq t \leq T} p_{t}$ be the minimal probability that all documents in a query are examined over time. We present the upper regret bound of \model{} as follows.
\begin{theorem}
\label{theorem}
Assume pairwise query-document feature vector $\xijt$ under query $q_t$, where $(i, j) \in [L_t]^2$ and $t \in [T]$, satisfies Proposition 1. With  $\delta_1 \in (0, \frac{1}{2})$, $\delta_2 \in (0, \frac{1}{2})$, $\beta \in (0, \frac{1}{2})$, the $T$-step regret of \model{} is upper bounded by:
\begin{align*}
    R_T 
     \leq& R^{\prime} + (1 - \delta_1)(1 - \delta_2) {p^*}^{-2}\left( 2adL_{\max}\log(1 + \frac{o_{\max}Tu^2}{2d\lambda}) + aw\right)^2
\end{align*}
where $R^{\prime} = \tp L_{\max}^2 + (T - t')\big(\delta_2L_{\max}^2 - (1- \delta_2)\delta_1 L_{\max}^2\big)$, with $\tp$ and $a$ defined in Lemma \ref{lemma:uncertain}, and $w = \sum\nolimits_{s=\tp}^T ({(L_{\max}^2 - 2L_{\max})u^2 }/({\lambda_{\min}(\bM_s)})$, and $L_{\max}$ representing the maximum number of document associated to the same query over time.
By choosing $\delta_1 = \delta_2 = 1/T$, we have the expected regret at most $R_T \leq O(d\log^4(T))$.
\end{theorem}

\begin{hproof}
The detailed proof is provided in the appendix. Here, we only provide the key ideas behind our regret analysis.
The regret is first decomposed into two parts: $R^\prime$ represents the regret when either $E_t$ or Lemma~\ref{lemma:uncertain} does not hold, in which the regret is out of our control, and we use the maximum number of pairs associated to a query over time, $L_{\text{max}}$ to compute the regret. The second part corresponds to the cases when both events happen. Then, the instantaneous regret at round $t$ can be bounded by
\begin{align}
    r_t = \mathbb{E} \big[K(\tau_t, \tau_t^*)\big] = \sum\nolimits_{i=1}^{d_t}\mathbb{E}\big[\frac{(N_i^t + 1)N_i^t}{2}\big] \leq \mathbb{E}\big[\frac{N_t(N_t + 1)}{2}\big]
\end{align}
where $N_i^t$ denotes the number of uncertain rank orders in block $\mathcal{B}_i^t$ at round $t$, and $N_t$ denotes the total number of uncertain rank orders.
From the last inequality, it follows that in the worst case where the $N_t$ uncertain rank orders are placed into the same block and thus generate at most $({N_t^2 + N_t})/{2}$ mis-ordered pairs with random shuffling. This is because based on the blocks created by \model{}, with $N_t$ uncertain rank orders in one block, this block can at most have $N_t + 1$ documents. Then, the cumulative number of mis-ordered pairs can be bounded by the probability of observing uncertain rank orders in each round, which shrinks rapidly with more observations over time.
\end{hproof}

\begin{remark}[1] 
By choosing $\delta = 1/T$, the theorem shows the expected regret increases at a rate of $\mathcal{O}({\log^4{T}})$. In this analysis, we provide a gap-dependent regret upper bound of \model{}, where the gap $\Delta_{\min}$ characterizes the hardness of sorting the $L_t$ candidate documents at round $t$. As the matrix $M_t$ only contains information from observed document pairs, we adopt the probability of a ranked list is fully observed to tackle with the partial feedback \cite{kveton2015combinatorial, kveton2015tight}, which is a constant independent of $T$.
\end{remark}

\begin{remark}[2]
Our regret is defined over the number of mis-ordered pairs, which is the \emph{first} pairwise regret analysis for an OL2R algorithm, to the best of our knowledge. As we discussed before, existing OL2R algorithms optimize their own metrics, which can hardly link to any conventional rank metrics. As shown in \cite{Wang2018Lambdaloss}, most classical ranking evaluation metrics, such as ARP and NDCG, are based on pairwise document comparisons. Our regret analysis of \model{} connects its theoretical property with such metrics, which has been later confirmed in our empirical evaluations.   
\end{remark}

%% file: experiment.tex
\section{Experiments}

\begin{figure*}[t]
  \centering
  \begin{subfigure}[b]{\textwidth}
    \centering
    \includegraphics[width=\linewidth]{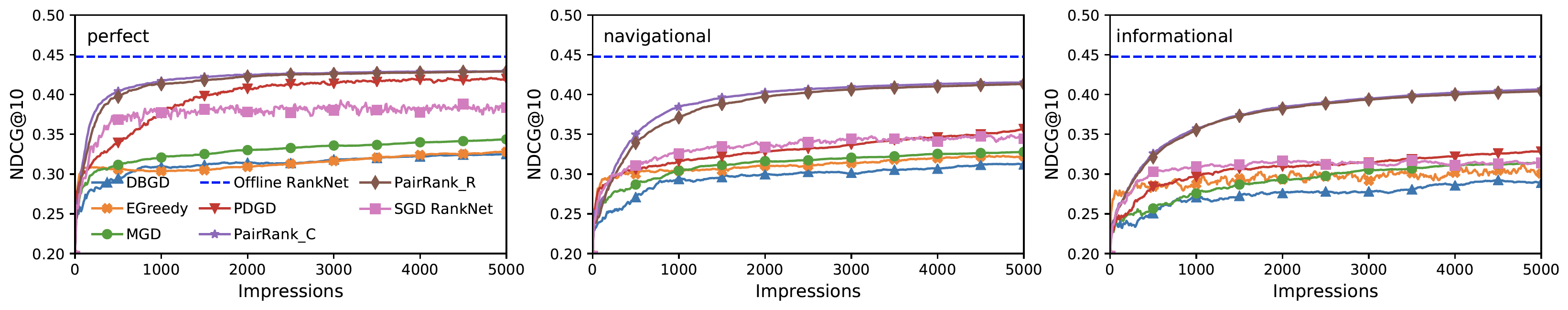}
    \caption{Offline performance (NDCG@10) on the MSLR-WEB10K dataset.}
    \label{fig:offline_WEB10K}
  \end{subfigure}
  \begin{subfigure}[b]{\textwidth}
    \centering
    \includegraphics[width=\linewidth]{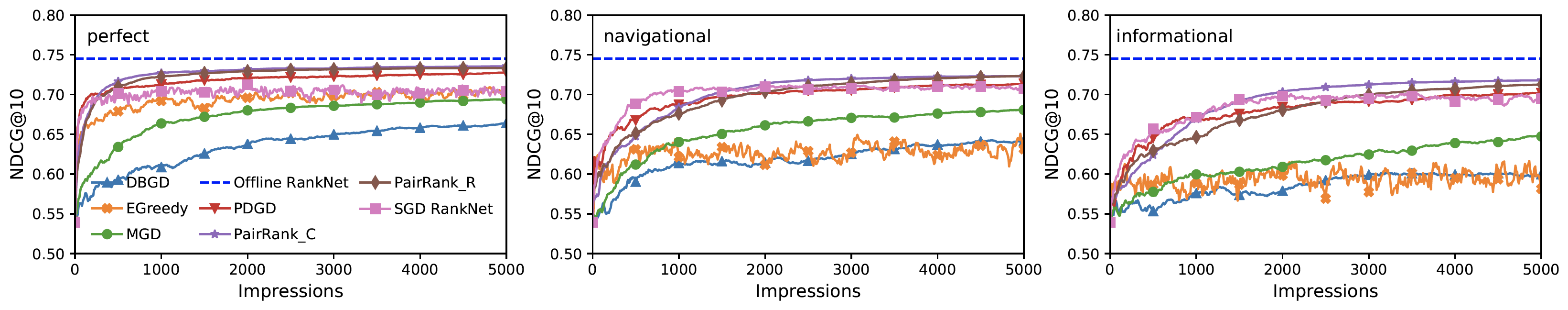}
    \caption{Offline performance (NDCG@10) on the Yahoo! dataset.}
    \vspace{-2mm}
    \label{fig:offline_Yahoo}
  \end{subfigure}
  \caption{Offline ranking performance on two different datasets under three different click models.}
  \label{fig:result}
  \vspace{-4mm}
\end{figure*}

In this section, we empirically compare our proposed \model{} with an extensive list of state-of-the-art OL2R algorithms on two public learning to rank benchmark datasets. Both quantitative and qualitative evaluations are performed to examine our proposed solution, especially its advantages over existing OL2R solutions in online learning efficiency. 

\noindent{\bf Reproducibility}
Our entire codebase, baselines, analysis, and experiments can be found on
Github~\footnote{https://github.com/yilingjia/PairRank}.

\subsection{Experiment Setup}

\textbf{Datasets.} We experiment on two publicly available learning to rank datasets: 1) Yahoo! Learning to Rank Challenge dataset \cite{chapelle2011yahoo}, which consists of 292,921 queries and 709,877 documents represented by 700 ranking features; and 2) MSLR-WEB10K \cite{qin2013introducing}, which contains 30,000 queries, each having 125 assessed documents on average, and is represented by 136 ranking features. Both datasets are labeled with a five-grade relevance scale: from not relevant (0) to perfectly relevant (4). We followed the train/test/validation split provided in the datasets.

\begin{table}
  \caption{Configuration of simulated click models.}
  \vspace{-2mm}
  \label{table:click}
  \centering
  \begin{tabular}{cccc|ccc}
    \hline
                & \multicolumn{3}{c}{Click Probability} & \multicolumn{3}{c}{Stop Probability} \\
Relevance Grade & 0           & 1          & 2          & 0          & 1          & 2          \\ \hline
Perfect         & 0.0         & 0.5        & 1.0        & 0.0        & 0.0        & 0.0        \\
Navigational    & 0.05        & 0.5        & 0.95       & 0.2        & 0.5        & 0.9        \\
Informational   & 0.4         & 0.7        & 0.9        & 0.1        & 0.3        & 0.5        \\ \hline
\end{tabular}
\vspace{-5mm}
\end{table}

\noindent\textbf{User Interaction Simulation.} We simulate user behavior via the standard setup for OL2R evaluations \cite{oosterhuis2018differentiable, wang2019variance} to make our reported results directly comparable to those in literature. First, at each time $t$, a query is uniformly sampled from the training set for result serving. Then, the model determines the ranked list and returns it to the users. The interaction between the user and the list is then simulated with a dependent click model~\cite{guo2009efficient}, which assumes that the user will sequentially examine the list and make a click decision on each document. At each position, the user decides whether to click on the document or not, modeled as a probability conditioned on the document's relevance label, e.g, $\mathbb{P}(click = 1 | \text{relevance grade})$. After the click, the user might stop due to his/her satisfaction with the result or continue to examine more results. The stop probability after a click is modeled as $\mathbb{P}(stop = 1 |click = 1, \text{relevance grade})$. If there is no click, the user will continue examining the next position. We employ three different click model configurations to represent three different types of users, for which the details are shown in Table \ref{table:click}. Basically, we have the \textit{perfect} users, who click on all relevant documents and do not stop browsing until the last returned document; the \textit{navigational} users, who are very likely to click on the first highly relevant document and stop there; and the \textit{informational} users, who tend to examine more documents, but sometimes click on irrelevant ones, and thus contribute a significant amount of noise in their click feedback.

\noindent\textbf{Baselines.} As our \model{} learns a parametric ranker, we focus our comparison against existing parametric OL2R solutions, which are known to be more generalizable and powerful than those estimating pointwise query-document relevance \cite{radlinski2008learning,kveton2015cascading,zoghi2017online,lattimore2018toprank}. We list the OL2R solutions used for our empirical comparisons below.

\noindent $\bullet$ \textbf{DBGD} \cite{yue2009interactively}: As detailed in our related work discussion, DBGD uniformly samples an exploratory direction from the entire model space for exploration and model update.

\noindent $\bullet$ \textbf{MGD} \cite{schuth2016multileave}: It improves DBGD by sampling multiple directions and compares them via a multileaving test. If there is a tie, the model updates towards the mean of all winners.

\noindent $\bullet$  \textbf{PDGD} \cite{oosterhuis2018differentiable}: 
PDGD samples the next ranked document from a Plackett-Luce model and estimates the gradient from the inferred pairwise result preferences.

\noindent $\bullet$  \textbf{$\epsilon$-Greedy} \cite{hofmann2013balancing}: It randomly samples an unranked document with probability $\epsilon$ or selects the next best document based on the currently learned RankNet with probability $1 - \epsilon$, at each position.

\noindent $\bullet$ \textbf{SGD RankNet}: It estimates a single layer RankNet with stochastic gradient descent using the same click preference inference method in \model{}. At each round, it presents the best ranking based on the currently learned RankNet model.

\noindent $\bullet$ \textbf{PairRank}: We compare two shuffling strategies within each partition: 1) random shuffling the documents, denoted as \model{}\_R, and 2) shuffling with respect to the uncertain rank orders, while preserving the certain rank orders within each partition, denoted as \model{}\_C (as illustrated in Figure \ref{fig:model}).

\subsection{Experiment results}

\subsubsection{Offline Performance}
In offline evaluation, while the algorithm is learning, we evaluate the learned rankers on a separate testing set using its ground-truth relevance labels. We use Normalized Discounted Cumulative Gain at 10 (NDCG@10) to assess different algorithms' ranking quality. We compare all algorithms over 3 click models and 2 datasets. We set the hyper-parameters, such as learning rate, in DBGD, MGD, and PDGD according to their original papers. For SGD RankNet, $\epsilon$-Greedy, and our proposed \model{}, we set the hyper-parameters via the best performance on the validation data for each dataset. 
We fix the total number of iterations $T$ to 5000. We execute the experiments
20 times with different random seeds and report the averaged result in Figure~\ref{fig:result}. For comparison purposes, we also include the performance of an offline RankNet trained on the ground-truth relevance labels on the training set, whose performance can be considered as an upper bound in our evaluation.

\begin{figure*}[t]
  \centering
  \begin{subfigure}[b]{\textwidth}
    \centering
    \includegraphics[width=\linewidth]{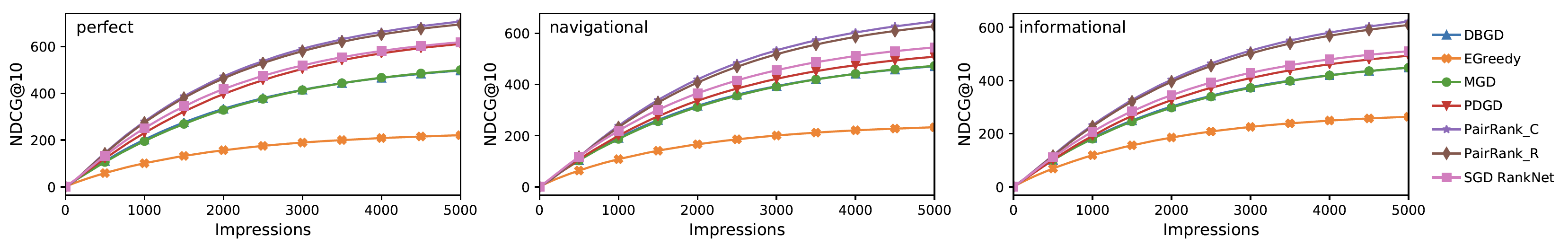}
    \vspace{-4mm}
    \caption{Online performance (cNDCG@10) on the MSLR-WEB10K dataset.}
    \label{fig:online_WEB10K}
  \end{subfigure}
  \begin{subfigure}[b]{\textwidth}
    \centering
    \includegraphics[width=\linewidth]{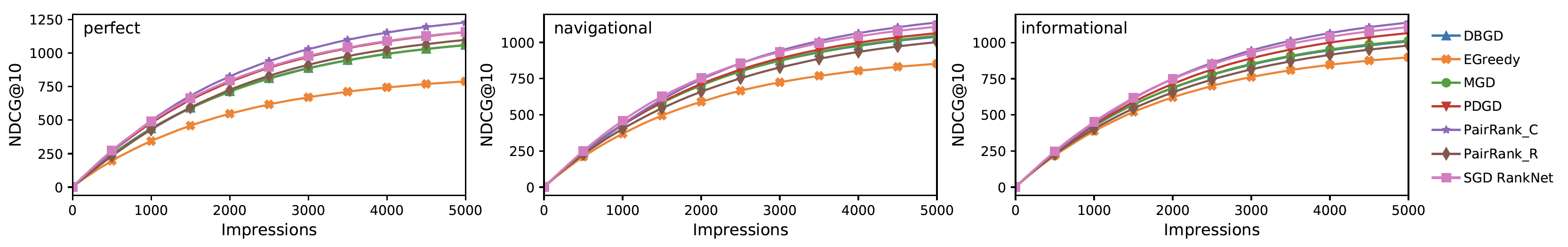}
    \vspace{-4mm}
    \caption{Online performance (cNDCG@10) on the Yahoo! dataset.}
    \label{fig:online_Yahoo}
  \end{subfigure}
  \vspace{-6mm}
  \caption{Online ranking performance on two different datasets under three different click models.}
  \label{fig:online}
  \vspace{-2mm}
\end{figure*}

We can clearly observe that \model{} achieved significant improvement over all the baselines. Across the two datasets under different click models, DBGD and MGD performed worse than other OL2R solutions. This is consistent with previous studies' findings: DBGD and MGD depend on interleave test to determine the update direction across rankers. But such model-level feedback is hard to directly inform the optimization of any rank-based metric (e.g., NDCG). PDGD consistently outperformed DBGD-type solutions under different click models and datasets. However, its document sampling-based exploration limits its learning efficiency, especially when users only examine a small number of documents (e.g., the navigational users).
$\epsilon$-Greedy seriously suffered from its random exploration, which is independent of how the current ranker performs. Oftentimes as shown in Figure~\ref{fig:offline_WEB10K}, its performance is even worse than MGD, although $\epsilon$-Greedy can directly optimize the pairwise ranking loss.  
One interesting observation is that SGD RankNet performs comparably to PDGD. Exploration in SGD RankNet is implicitly achieved via stochastic gradient descent. However, because this passive exploration is subject to the given queries and noise in user feedback, its performance in online result serving is hard to predict. 
We attribute \model{}'s fast convergence to its uncertainty based exploration: it only explores when its estimation on a pair of documents is uncertain. As proved in our regret analysis, the number of such pairs shrinks at a logarithmic rate, such that more and more documents are presented in their correct ranking positions as \model{} learns from user feedback. This conclusion is further backed by the comparison between \model{}\_R and \model{}\_C: by respecting the certain rank orders within a partition, \model{}\_C further improves result ranking quality. When click noise is small, i.e., from perfect users, \model{} converges to its offline counterpart with only a few thousand impressions.

\subsubsection{Online Performance}

In OL2R, in addition to the offline evaluation, a model's ranking performance during online optimization should also be evaluated, as it reflects user experience. Sacrificing user experience for model training will compromise the goal of OL2R. We adopt the cumulative Normalized Discounted Cumulative Gain~\cite{oosterhuis2018differentiable} to assess models' online performance. For $T$ rounds, the cumulative NDCG (cNDCG) is calculated as
\small
\begin{equation*}
    \text{cNDCG} = \sum\nolimits_{t=1}^T \text{NDCG}(\tau_t) \cdot \gamma^{(t-1)},
\end{equation*}
\normalsize
which computes the expected NDCG reward a user receives with a probability $\gamma$ that he/she stops searching after each query~\cite{oosterhuis2018differentiable}. Following the previous work~\cite{oosterhuis2018differentiable, wang2019variance, wang2018efficient}, we set $\gamma = 0.9995$.

Figure~\ref{fig:online} shows the online performance of \model{} and all the other baselines. It is clear to observe that directly adding model-independent exploration, e.g., $\epsilon$-greedy, strongly hurts a model's online ranking performance, and therefore hurts user experience. 
Compared to PDGD, SGD RankNet showed consistently better performance, especially under the navigational and informational click models. We attribute this difference to the exploration strategy inside PDGD: though SGD RankNet has limited exploration and focuses more on exploiting the learned model in ranking the documents, PDGD's sampling-based exploration might introduce unwanted distortion in the ranked results, especially at the early stage of online learning. 

We should note in cumulative NDCG, the earlier stages play a much more important role due to the strong shrinking effect of $\gamma$. Our proposed \model{} demonstrated significant improvements over all the baselines.
Such improvement indicates the effectiveness of our uncertainty based exploration, which only explores when the ranker's pairwise estimation is uncertain. 
We can also observe the difference between PairRank\_C and PairRank\_R in this online evaluation. This is because PairRank\_C preserves the certain rank order within blocks, which further eliminates unnecessary exploration (random shuffling) in result serving.

\subsubsection{Closeness to the \emph{Ideal} Ranker}

\begin{figure}[t]
    \centering
    \begin{subfigure}[b]{0.5\textwidth}
    \includegraphics[width=0.95\textwidth]{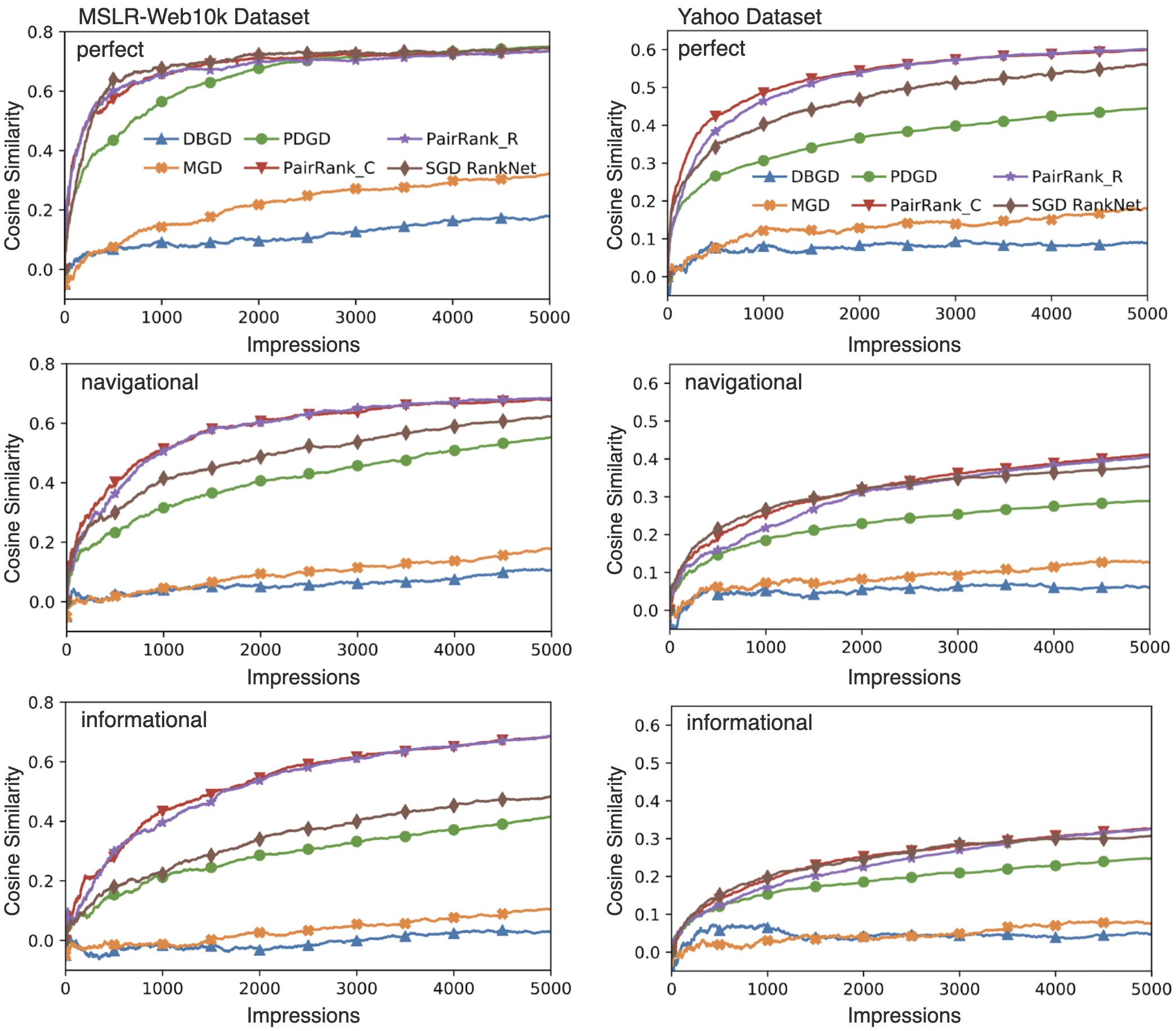}
    \end{subfigure}
    \vspace{-4mm}
    \caption{Cosine similarity between offline RankNet model and online models.}
    \label{fig:cosine}
    \vspace{-4mm}
\end{figure}

In this experiment, we view the offline trained RankNet model on the complete annotated relevance as the ideal model, denoted as $w^*$, and compare the cosine similarities between $w^*$ and the online estimated models in Figure~\ref{fig:cosine}. We can observe that all the pairwise models push the learned ranker closer to the ideal one with more iterations, while \model{} converges faster and closer to $w^*$. Apparently, the rankers obtained by DBGD and MGD are quite different from the ideal model. This confirms our earlier analysis that DBGD-type ranker update can hardly link to any rank-based metric optimization, and thus it becomes less related to the offline learned ranker.  
The difference in objective function can also explain why SGD RankNet converges faster than PDGD, as PDGD adopts the pairwise Plackett-Luce model while SGD RankNet has the same objective function as the offline model.

\subsubsection{Zoom into \model{}}

\begin{figure}[t]
    \centering
    \begin{subfigure}[b]{0.235\textwidth}
    \includegraphics[width=\textwidth]{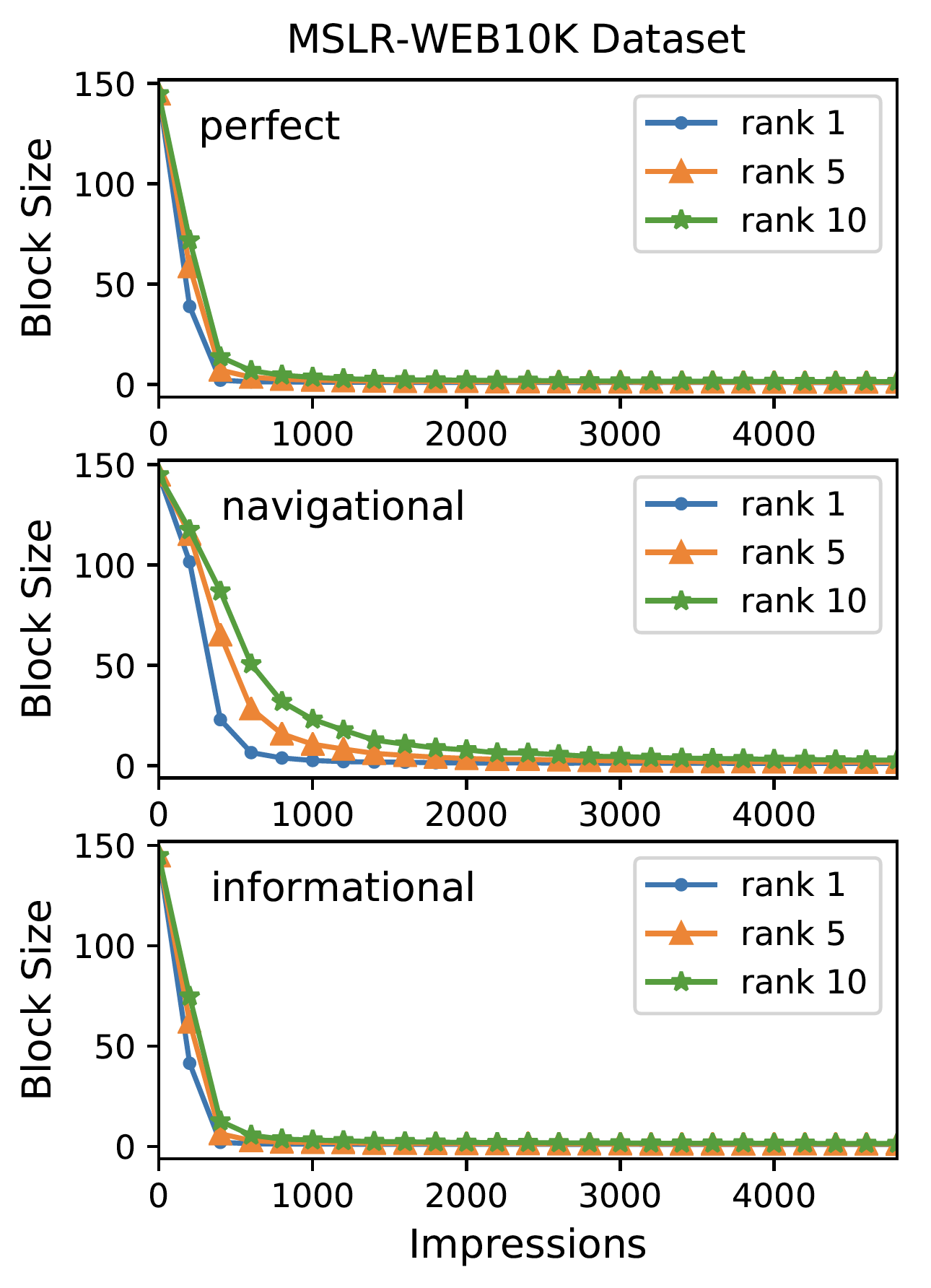}
    \caption{MSLR-WEB10K Dataset}
    \label{fig:block_size_web10k}
    \end{subfigure}
    \begin{subfigure}[b]{0.23\textwidth}
    \includegraphics[width=\textwidth]{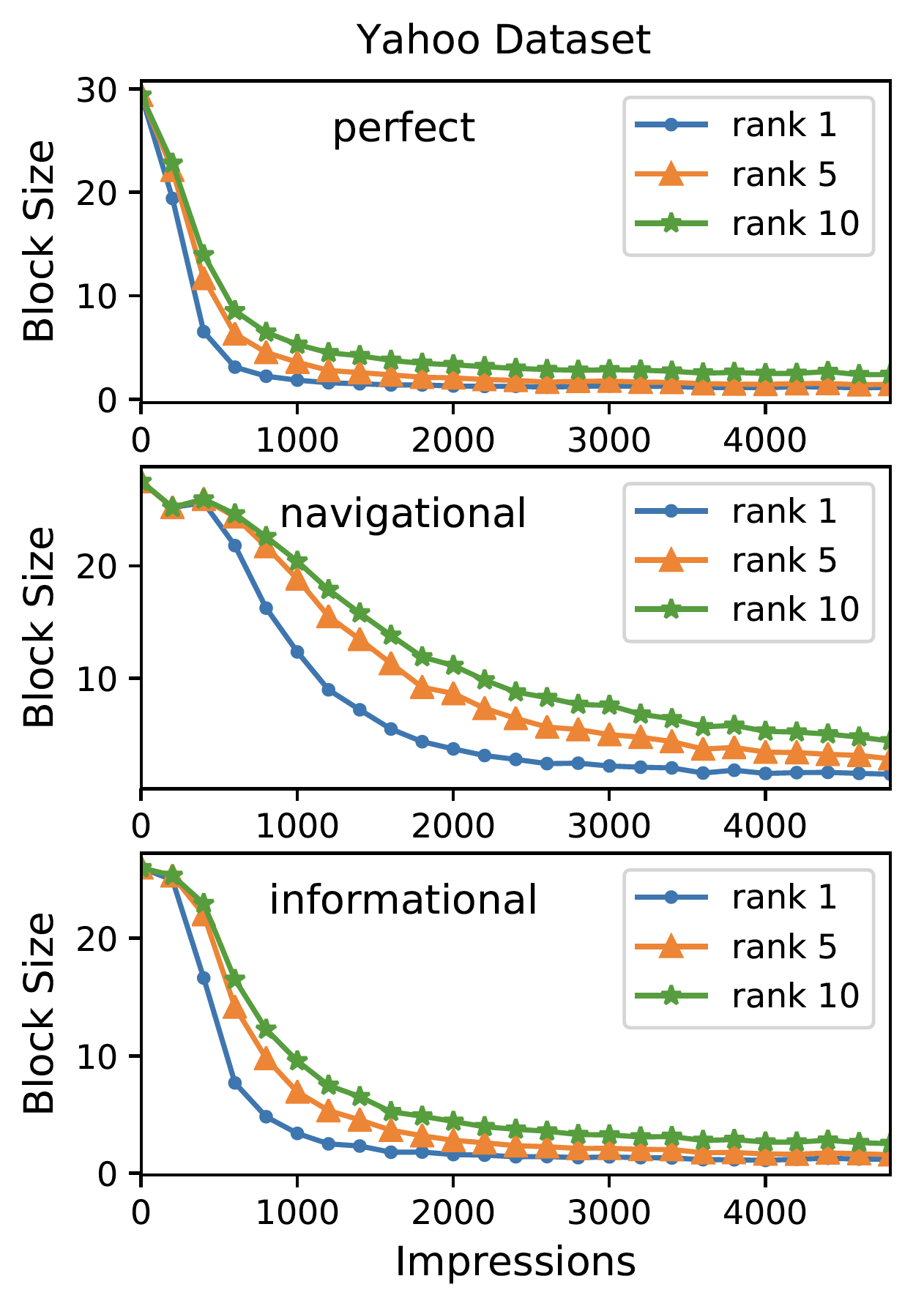}
    \caption{Yahoo Dataset}
    \label{fig:block_size_yahoo}
    \end{subfigure}
    \vspace{-4mm}
    \caption{The size of blocks at top ranks.}
    \label{fig:block_size}
    \vspace{-4mm}
\end{figure}

To further verify the effectiveness of the exploration strategy devised in \model{}, we zoom into the trace of its block size across queries during the online model update. As \model{} uses random shuffling within each block for exploration, a smaller block size, especially at the top-ranked positions, is preferred to reduce regret. Figure~\ref{fig:block_size} shows the size of document blocks at rank position 1, 5 and 10. 

\begin{figure}[t]
    \centering
    \begin{subfigure}[b]{0.23\textwidth}
    \includegraphics[width=\textwidth]{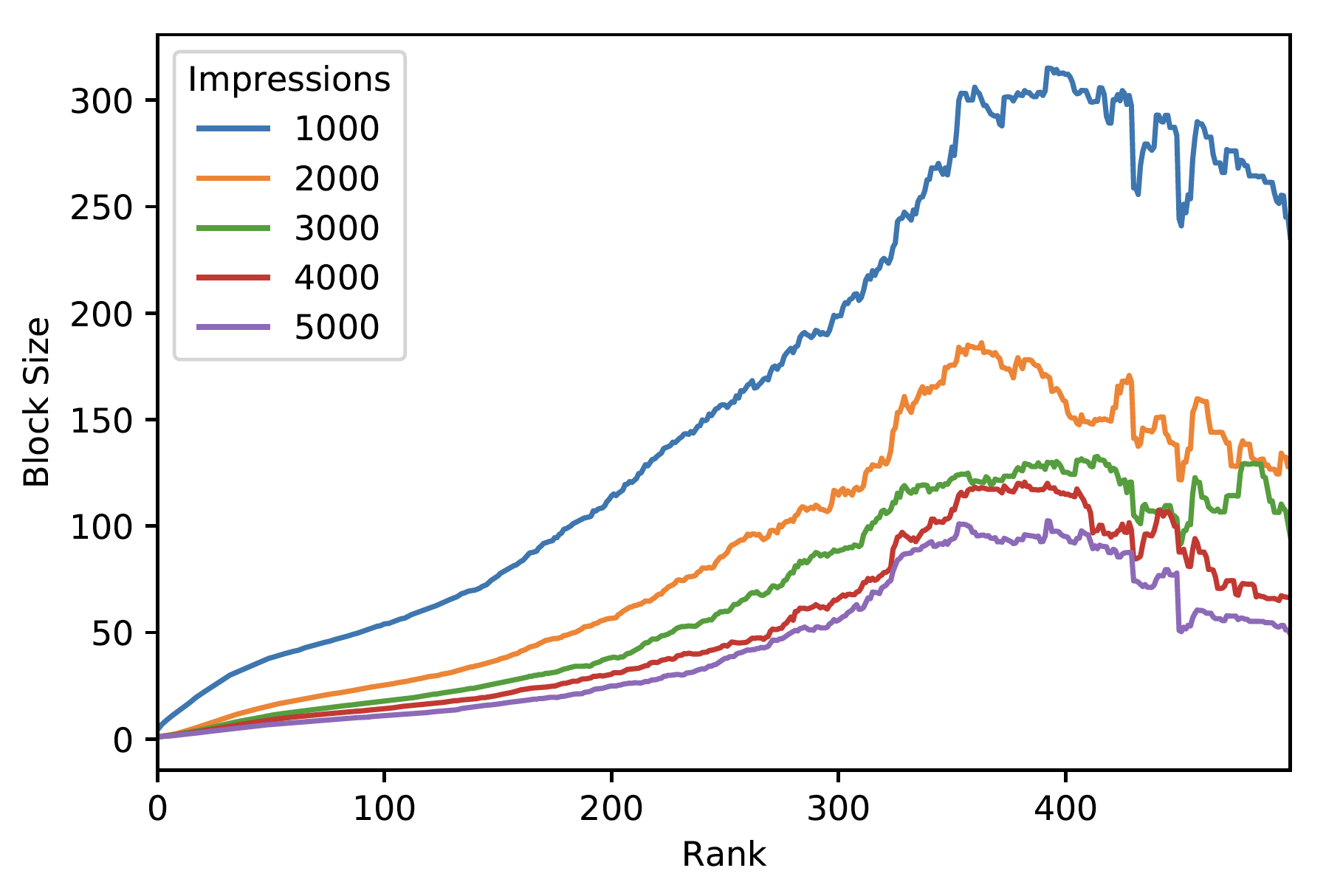}
    \caption{MSLR-WEB10K with informational click model.}
    \label{fig:block_size_inf}
    \end{subfigure}
    \begin{subfigure}[b]{0.23\textwidth}
    \includegraphics[width=\textwidth]{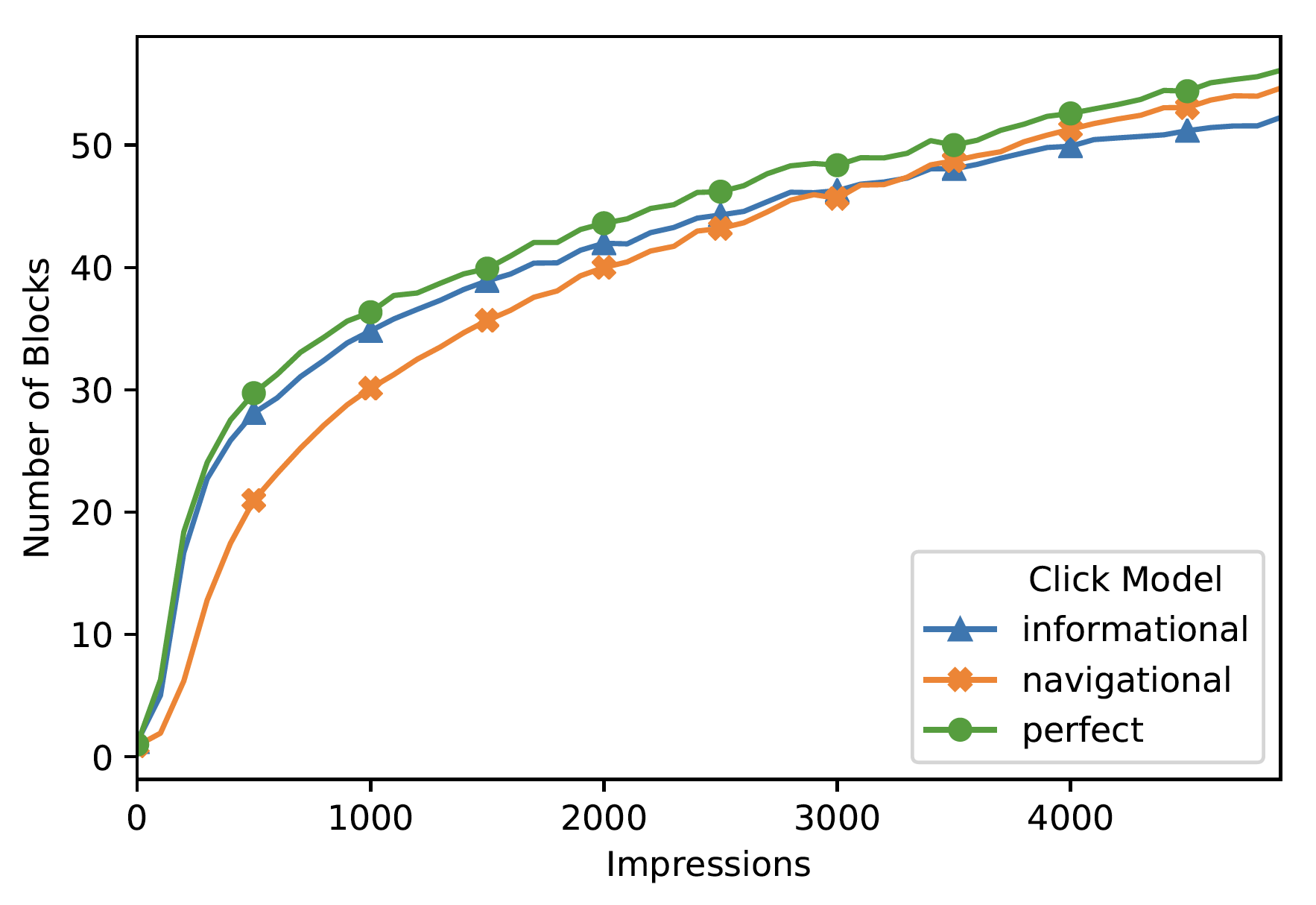}
    \caption{Number of blocks identified on MSLR-WEB10K dataset}
    \label{fig:number_of_block}
    \end{subfigure}
    \caption{Qualitative analysis of blocks in \model{}}
    \label{fig:block_analysis}
\end{figure}

First, we can clearly observe that after hundred rounds of interactions, the sizes of blocks quickly converge to 1, especially at the top-ranked positions. This confirms our theoretical analysis about \model{}'s block convergence. And by comparing the results across different click models, we can observe that the block size converges slower under the navigational click model. Similar trends can be observed in Figure~\ref{fig:number_of_block}. In this figure, the number of blocks is calculated by averaging in every 200 iterations to reduce variance. We can observe that at the early stages, \model{} under the navigational click model has fewer blocks (hence more documents in one block), which indicates a higher uncertainty in model estimation. The key reason is that much fewer clicks can be observed in each round under the navigational click model, as the stop probability is much higher, i.e., stronger position bias. As a result, more interactions are needed to improve model estimation. For the same reason, in Figure~\ref{fig:block_size}, the block size at rank position 10 shrinks much slower than that at other positions also suggests position bias is the main factor that slows down the learning of \model{}. 

In Figure~\ref{fig:block_size_inf}, we show the block size at each rank position in \model{} under the informational click model at different time points. One interesting finding is the blocks in the middle ranks, i.e., rank 300 - 400, tend to have more documents, while for the ranks at both the top and the bottom, the corresponding blocks tend to have fewer documents. We attribute it to the pairwise learning in \model{}, where the uncertainty is calculated on the pairwise preferences, and thus it is easy to identify the document pairs with greater differences.

%% file: supplementary.tex
\section{Preliminaries}
In this section, we present some basic definitions and inequalities for later use.
The first inequality is about the Lipschitz continuity of the logistic function.
\begin{lemma} For the logistic function $\sigma(x) = {1}/{(1 + \exp(-x))}$, we have:
$|\sigma(x) - \sigma(y)| < k_{\mu} |x - y|$
where $k_{\mu} = 1/4$.
\label{lemma:lipt}
\end{lemma}

The second inequality is the self-normalized bound for martingales, adopted from Lemma 9 in \cite{abbasi2011improved}.

\begin{lemma}[Self-normalized bound for martingales, Lemma 9 in \cite{abbasi2011improved}]
\label{lemma:martigale}
Let $t$ be a stopping time with respect to the filtration $\{F_\tau\}^\infty_{\tau=0}$. Because $\epsilon_{ij}^s$ is $\frac{1}{2}$-sub-Gaussian, for $\delta > 0$, with probability $1-\delta$,
\small
\begin{equation*}
    \left\| \sum\nolimits_{s=1}^{t-1}\sum\nolimits_{(i, j)\in \mathcal{G}_s^{ind}} \epsilon_{ij}^s\bx^s_{ij}\right\|_{\sub \bM_t^{-1}}^2 \leq ({1}/{2})\log({{\det(\bM_t)^{1/2}}/{(\delta \det(\lambda\mathbf{I})^{1/2})}})
\end{equation*}
\normalsize
\end{lemma}

\section{Proof of \model{}'s confidence interval: Lemma~\ref{lemma:cb}}


\subsection{Model learning for PairRank}
\label{sec:modellearning}

As we introduced in Section 3.2, we adopt a single layer RankNet model with sigmoid activation function as our pairwise ranker. The objective function for our ranker defined in Eq~(2) is the cross-entropy loss between the predicted pairwise preference distribution and the inferred distribution from user feedback. Given this objective function is log-convex with respect to $\btheta$, its solution $\tilde{\btheta}_t$ is unique under the following estimating equation at each round $t$,
\small
\begin{align}
\label{eqn:gradient}
    \sum\nolimits_{s=1}^{t-1}\sum\nolimits_{\scriptscriptstyle (m, n)\in\mathcal{G}_{s}^{ind}} \left(\sigma({\xmns}^\top\btheta) - \ymns \right)\xmns+ \lambda\btheta = 0
\end{align}
\normalsize
Let $g_t(\btheta) = \sum_{\scriptscriptstyle s=1}^{\scriptscriptstyle t-1}\sum_{(m, n) \in \mathcal{G}_s^{ind}}\sigma({\xmns}^\top\btheta)\xmns + \lambda\btheta$ be the invertible function such that the estimated parameter $\tilde{\btheta}_t$ satisfies $g_t(\tilde{\btheta}_t) = \sum_{s=1}^{t-1}\sum_{(m, n)\in\mathcal{G}_{s}^{ind}}\ymns\xmns$. Since $\tilde{\btheta}_t$ might be outside of the admissible set of parameter $\Theta$ that $\btheta \in \Theta$ (e.g., $\Vert\btheta\Vert_2 \leq Q$), the estimation can be projected into $\Theta$ to obtain $\hat{\btheta}_t$:
\small
\begin{align}
\label{eq:opt-solution}
    \hat{\btheta}_t 
    &{=} \argmin_{\btheta \in \Theta} \left\| g_t(\btheta) {-} g_t(\tilde{\btheta}_t)\right\|_{\scriptscriptstyle \bM_t^{-1}} \\
    &{=} \argmin_{\btheta \in \Theta} \left\| g_t(\btheta) {-} \sum_{\scriptscriptstyle s=1}^{\scriptscriptstyle t-1}\sum_{\scriptscriptstyle (m, n)\in\mathcal{G}_{s}^{ind}}\ymns\xmns\right\|_{\scriptscriptstyle \bM_{t}^{-1}} \nonumber
\end{align}
\normalsize
where $\bM_t$ is defined as $\bM_t = \sum_{s =1}^{t-1}\sum_{(m, n) \in \mathcal{G}_{s}^{ind}}\xmns{\xmns}^\top + \lambda \mathbf{I}$. To summarize, $\tilde{\btheta}_t$ is the solution of Eq~(4), and $\hat{\btheta}_t$ is the estimated model which is generated by projecting $\tilde{\btheta}_t$ onto $\Theta$.

\subsection{Confidence interval analysis}
\label{sec:cb}
Now we present detailed proof of the confidence interval in Lemma~\ref{lemma:cb}, which is based on Proposition 1 in~\cite{filippi2010parametric}.

At round $t$, for any document pair $\bx^t_i$ and $\bx^t_j$, the estimation error of the pairwise preference in \model{} is defined as $|\sigma({\xijt}^\top\btheta^*) - \sigma({\xijt}^\top\hat{\btheta}_t)|$, which is based on the model $\hat\btheta_t$ learned from the observations until last round. According to Lemma~\ref{lemma:lipt}, we have $|\sigma({\xijt}^\top\btheta^*) - \sigma({\xijt}^\top\hat{\btheta}_t)| \leq k_{\mu}|{\xijt}^\top \btheta^* - {\xijt}^\top\hat{\btheta}_t|$. As logistic function $\sigma(\cdot)$ is continuously differentiable, $\nabla g_t$ is continuous. Hence, according to the Fundamental Theorem of Calculus, we have $g_t(\btheta^*) - g_t(\hat{\btheta}_t) = \bG_t(\btheta^* - \hat{\btheta}_t)$, where $\bG_t = \int _0^1\nabla g_t \left(l\btheta^* + (1 - l)\hat{\btheta}_t\right) dl.$ Therefore, $\nabla g_t(\btheta) = \sum_{s=1}^{t-1}\sum_{(m, n)\in \mathcal{G}_s^{ind}} \dot\sigma({\xmns}^\top\btheta) {\xmns}{\xmns}^\top + \lambda\mathbf{I}$, where $\dot{\sigma}$ is the first order derivative of $\sigma$. As $c_{\mu} = \inf_{\btheta \in \bTheta} \dot{\sigma}(\bx^\top\btheta)$, it is easy to verify that $c_{\mu} \leq 1/4$. Thus, we can conclude that $\bG_t \succeq c_{\mu}\bM_t$. Accordingly, we have the following inequality,
\small
\begin{align*}
    &\left|\sigma({\xijt}^\top\btheta^*) - \sigma({\xijt}^\top\hat{\btheta}_t) \right| \\
    \leq& k_{\mu} \left| {\xijt}^\top \mathbf{G}_t^{-1} \left(g_t(\btheta^*) - g_t(\hat{\btheta}_t)\right) \right| \leq k_{\mu} \Vert\xijt\Vert_{\scriptscriptstyle \mathbf{G}_t^{-1}}\left\| g_t(\btheta^*) - g_t(\hat{\btheta}_t) \right\|_{\scriptscriptstyle \mathbf{G}_t^{-1}} \\
    \leq& ({2k_{\mu}}/{c_{\mu}}) \Vert\xijt\Vert_{\scriptscriptstyle\bM_t^{-1}}\left\| g_t(\btheta^*) - g_t(\tilde{\btheta}_t) \right\|_{\scriptscriptstyle\bM_t^{-1}} 
\end{align*}
\normalsize
where the first and second inequalities stand as $\mathbf{G}_t$ and $\mathbf{G}_t^{-1}$ are positive definite. The third inequality stands as $\mathbf{G}_t \succeq c_{\mu}\bM_t$, which implies that $\mathbf{G}_t^{-1} \preceq c_{\mu}^{-1} \bM_t^{-1}$ and $\Vert\bx\Vert_{\sub \mathbf{G}_t^{-1}} \leq \Vert\bx\Vert_{\sub \bM_t^{-1}}/{\sqrt{c_{\mu}}}$ hold for any $\bx \in \mathbb{R}^d$. The last inequality stands as $\btheta^* \in \Theta$, and $\hat{\btheta}_t$ is the optimum solution for Eq \eqref{eq:opt-solution} at round $t$ within $\Theta$. 

Based on the definition of $\tilde \btheta_t$ and the assumption on the noisy feedback that $y^t = \sigma({\bx_t}^\top\btheta^*) + \epsilon^t$, where $\epsilon^t$ is the noise in user feedback defined in Section 3.2, we have 
\small
\begin{align*}
    &g_t(\tilde{\btheta}_t) - g_t(\btheta^*) \\
    =& \sum\nolimits_{s=1}^{t-1}\sum\nolimits_{(m, n)\in\mathcal{G}_{s}^{ind}}(\ymns -  \sigma({\xmns}^\top\btheta^*)) \xmns - \lambda\btheta^* \\
    =& \sum\nolimits_{s=1}^{t-1}\sum\nolimits_{(m, n)\in \mathcal{G}_s^{ind}} \epsilon_{mn}^s\xmns - \lambda\btheta^* = S_t - \lambda\btheta^*.
\end{align*}
\normalsize
where we define $S_t = \sum_{s=1}^{t-1}\sum_{(m, n)\in\mathcal{G}_{s}^{ind}}\epsilon^s_{mn}\xmns$.

Therefore, the confidence interval of the estimated pairwise preference in \model{} can be derived as:
\small
\begin{align*}
    \left|\sigma(\xijt^\top\btheta^*) - \sigma(\xijt^\top\hat{\btheta}_t) \right| \leq & ({2k_{\mu}}/{c_{\mu}}) \Vert\xijt\Vert_{\bM_t^{-1}}\left\| S_t - \lambda\btheta^*\right\|_{\bM_{t}^{-1}} \\
    \leq & ({2k_{\mu}}/{c_{\mu}}) \Vert\xijt\Vert_{\bM_t^{-1}}\left(\left\| S_t\right\|_{\bM_t^{-1}} + \sqrt{\lambda}Q\right)
\end{align*}
\normalsize
where the second inequality is based on minimum eigenvalue $\lambda_{\min}(\bM_t) \geq \lambda$ and $\Vert\btheta\Vert_2 \leq Q$.

As $\epsilon_{mn}^t \sim R$-sub-Gaussian, according to Theorem 1 in~\cite{abbasi2011improved},
\small
\begin{equation*}
    \mathbb{P}\left[ \left\| S_t\right\|_{\bM_t^{-1}}^2 > 2R^2\log{\frac{\det(\bM_t)^{1/2}}{\delta \det(\lambda\mathbf{I})^{1/2}}}\right] \leq \delta
\end{equation*}
\normalsize

Therefore, with probability at least $1 - \delta_1$, we have
\begin{equation*}
    \left|\sigma({\xijt}^\top \hat{\btheta}_t) - \sigma({\xijt}^\top \btheta^*) \right| \leq \alpha_t\Vert\xijt\Vert_{\bM_t^{-1}}
\end{equation*}
with $\alpha_t = ({2k_{\mu}}/{c_{\mu}}) \Big(\sqrt{R^2\log{\frac{\det(\bM_t)}{\delta_1^2 \det(\lambda \mathbf{I})}}} + \sqrt{\lambda} Q\Big)$

\section{Proof of Theorem~\ref{theorem}}
In this section, we present the detailed proof of Theorem 1. We first prove Lemma~\ref{lemma:uncertain}, which provides an upper bound of the probability that an estimated pairwise preference is identified as uncertain. The blocks with uncertain rank orders will lead to regret in the ranked list due to the random shuffling based exploration strategy.

\subsection{Proof of Lemma~\ref{lemma:uncertain}}

\begin{proof}

In this proof, we will first provide an upper bound of the minimum eigenvalue of $\bM_t$, and then provide the detailed derivation of the probability's upper bound.

\noindent{\bf $\bullet~$Upper bound the minimum eigenvalue of $\bM_t$}.
As discussion in Lemma~\ref{lemma:matrix} , we assume that pairwise feature vectors are random vectors drawn from  distribution $v$. With $\bSigma = \mathbb{E}[\xmns\xmns^\top]$ as the second moment matrix, define $\bZ = \bSigma^\frac{-1}{2}\mathbf{X}$, where $\mathbf{X}$ is a random vector drawn from the same distribution $v$. Then $\bZ$ is isotropic, namely $\mathbb{E}[\bZ\bZ^\top] = \mathbf{I}_d$. 

Define $\bU = \sum_{s=1}^{t-1}\sum_{(m, n) \in \mathcal{G}_s^{ind}} \bZ_{mn}^s{\bZ_{mn}^s}^\top = \Sigma^\frac{-1}{2}\bar{\bM}_t\Sigma^\frac{-1}{2}$, where $\bar{\bM}_t = \sum_{s=1}^{t-1}\sum_{(m, n)  \in \mathcal{G}_s^{ind}}\xmns\xmns^\top = \bM_t - \lambda\mathbf{I}$. From Lemma~\ref{lemma:matrix}, we know that for any $l$, with probability at least $1 - 2\text{exp}(-C_2l^2)$,
$\lambda_{\text{min}}(\bU) \geq n - C_1\sigma^2\sqrt{nd} - \sigma^2l\sqrt{n}$, 
where $\sigma$ is the sub-Gaussian parameter of $\bZ$, which is upper-bounded by  $\Vert\bSigma^{-1/2}\Vert = \lambda_{\text{min}(\bSigma)}$, and $n = \sum_{s=1}^{t-1}|\mathcal{G}_s^{ind}|$, represents the number of observations so far. We thus can rewrite the above inequality which holds with probability $1 - \delta_2$ as
$\lambda_{\text{min}}(\bU) \geq n - \lambda_{\text{min}}^{-1}(\bSigma)(C_1\sqrt{nd} + l\sqrt{n})$.
We now bound the minimum eigenvalue of $\bar{\bM}_t$, as follows:
\small
\begin{align*}
    \lambda_{\text{min}}(\bar{\bM}_t) &= \min_{x\in \mathbb{B}^d}x^\top\bar{\bM}_t x = \min_{x\in \mathbb{B}^d}x^\top\bSigma^{1/2}\bU\bSigma^{1/2}x \geq \lambda_{\text{min}}(\bU)\min_{x\in \mathbb{B}^d}x^\top \bSigma x \\
    &= \lambda_{\text{min}}(\bU)\lambda_{\text{min}}(\bSigma) \geq \lambda_{\text{min}}(\bSigma)n - C_1\sqrt{nd} - C_2\sqrt{n\log(1/\delta_2)}
\end{align*}
\normalsize
According to Lemma~\ref{lemma:matrix}, for $t \geq t^\prime$, we have:
\small
\begin{align*}
    &\lambda_{\text{min}}(\bSigma)t - (c_1\sqrt{d} + c_2\sqrt{\log({1}/{\delta_2})} + abd\sqrt{{o_{\text{max}}u^2}/({d\lambda})})\sqrt{t} \\
    & - (ab\log({1}/{\delta_1^2}) + 4a\lambda Q^2 - \lambda) \geq 0
\end{align*}
As $n \geq t$, we have
\begin{equation}\label{eq:lambda_min}
    \lambda_{\text{min}}({\bM}_t)  \geq  \lambda_{\text{min}}(\bar{\bM}_t) + \lambda \geq abd\sqrt{\frac{o_{\text{max}}u^2}{d\lambda}} + ab\log(\frac{1}{\delta_1^2}) + 4a\lambda Q^2
\end{equation}
\normalsize

\noindent{\bf $\bullet~$Upper bound the probability of being uncertain rank order}.
Under event $E_t$, based on the definition of $\mathcal{E}_u^t$ in Section 3.2, we know that for any document $\bx^t_i$ and $\bx^t_j$ at round $t$, $(i, j) \in \mathcal{E}_u^t$ if and only if $\sigma({\xijt}^\top\hat{\btheta}_t) - \alpha_t\Vert\xijt\Vert_{\bM_t^{-1}} \leq \frac{1}{2}$ and $\sigma({\xjit}^\top\hat{\btheta}_t) - \alpha_t\Vert\xjit\Vert_{\bM_t^{-1}} \leq \frac{1}{2}$. 
For a logistic function, we know that $\sigma(s) = 1 - \sigma(-s)$. Therefore, let $CB_{ij}^t$ denote $\alpha_t\Vert\xijt\Vert_{\bM_t^{-1}}$, we can conclude that $(i, j) \in \mathcal{E}_u^t$ if and only if $|\sigma({\xijt}^\top\hat{\btheta}_t )- \frac{1}{2}| \leq CB_{ij}^t$; and accordingly, $(i, j) \in \mathcal{E}_c^t$, when $|\sigma({\xijt}^\top\hat{\btheta}_t) - \frac{1}{2}| > CB_{ij}^t$.  To further simplify our notations, in the following analysis, we use $\hat{\sigma}_t$ and $\sigma^*$ to present $\sigma({\xijt}^\top\hat{\btheta}_t)$ and $\sigma({\xijt}^\top\btheta^*)$ respectively. 
According to the discussion above, the probability that the estimated preference between document $\bx^t_i$ and $\bx^t_j$ to be in an uncertain rank order, e.g., $(i, j) \in \mathcal{E}_u^t$ can be upper bounded by:
\small
\begin{align*}
    & \mathbb{P}\big((i, j) \in \mathcal{E}_u^t\big) = \mathbb{P}\big(|\hat{\sigma}_t - {1}/{2}| \leq CB_{ij}^t\big) \nonumber \\
    \leq& \mathbb{P}\left(\left||\hat{\sigma}_t - \sigma^*| - |\sigma^* - {1}/{2}|\right| \leq CB_{ij}^t\right) 
    \leq \mathbb{P} \left(|\sigma^* - {1}/{2}| - |\hat{\sigma}_t - \sigma^*| \leq CB_{ij}^t\right) \\
    \leq& \mathbb{P} \left(\Delta_{\min} - |\hat{\sigma}_t - \sigma^*| \leq CB_{ij}^t\right). 
\end{align*}
\normalsize
where, the first inequality is based on the reverse triangle inequality. 
The last inequality is based on the definition of $\Delta_{\min}$. Based on Lemma~\ref{lemma:cb}, the above probability can be further bounded by
\small
\begin{align*}
    &\mathbb{P} \left(\Delta_{\min} - |\hat{\sigma}_t - \sigma^*| \leq CB_{ij}^t\right) = \mathbb{P}\left(|\hat{\sigma}_t - \sigma^*| \geq \Delta_{\min} - \alpha_t\Vert\xijt\Vert_{\sub \bM_t^{-1}}\right) \\
    \leq& \mathbb{P} \left(({2k_{\mu}}/{c_{\mu}} )||\xijt||_{\sub \bM_t^{-1}}
    \left(\left\| S_t\right\|_{\sub \bM_t^{-1}} + \sqrt{\lambda}Q\right) \geq \Delta_{\min} - \alpha_t\Vert\xijt\Vert_{\sub \bM_t^{-1}}\right) \\
    \leq& \mathbb{P}\left(\left\| S_t\right\|_{\sub \bM_t^{-1}} \geq \frac{c_{\mu} \Delta_{\min}}{2k_{\mu}||\xijt||_{\sub \bM_{t}^{-1}}} - \left(\sqrt{\frac{1}{4}\log{\frac{\det(\bM_t)}{\delta^2 \det(\lambda \mathbf{I})}}} + 2\sqrt{\lambda} Q\right)\right)
\end{align*}
\normalsize
For the RHS of the inequality inside the probability, we know that:
\small
\begin{align*}
&\left(\frac{c_{\mu}\Delta_{\min}}{2k_{\mu}||\xijt||_{\sub \bM_{t}^{-1}}}
\right)^2 - \left(\sqrt{R^2\log{\frac{\det(\bM_t)}{\delta^2 \det(\lambda \mathbf{I})}}} + 2\sqrt{\lambda} Q\right)^2 \\
\geq& {\lambda_{\min}(\bM_t)}/{a}  - R^2\log({\det(\bM_t)}/{(\delta_1^2\det(\lambda\mathbf{I}))})  \\
&- 4\lambda Q^2 - 4\sqrt{\lambda}QR\sqrt{\log({\det(\bM_t)}/{\det{\lambda\mathbf{I}}}) + \log({1}/{\delta_1^2})} \\
\geq& {\lambda_{\min}(\bM_t)}/{a} - b\left(\log({\det(\bM_t)}/{\det(\lambda\mathbf{I})})  + \log({1}/{\delta_1^2})\right)- 4\lambda Q^2\\
\geq & {\lambda_{\min}(\bM_t)}/{a} - bd\log(1 + \frac{o_{\max}tu^2}{d\lambda}) - b\log(\frac{1}{\delta_1^2}) - 4\lambda Q^2 \geq  0
\end{align*}
\normalsize
where the last inequality follows Eq~\eqref{eq:lambda_min}. Therefore, the probability could be upper bounded as:
\small
\begin{align*}
    &\mathbb{P} \left(\Delta_{\min} - |\hat{\sigma}_t - \sigma^*| \leq CB_{ij}^t\right) \\\leq
    &\mathbb{P}\left(\left\| S_t\right\|_{\sub \bM_t^{-1}}^2  {\geq} \left(\frac{c_{\mu} \Delta_{\min}}{2k_{\mu}\Vert\xijt\Vert_{\sub \bM_{t}^{-1}}} - \Big(\sqrt{R^2\log{\frac{\det(\bM_t)}{\delta \det(\lambda \mathbf{I})}}} + 2\sqrt{\lambda} Q\Big)\right)^2\right)\\
    \leq &  \mathbb{P}\left(\left\| S_t\right\|_{\sub \bM_t^{-1}}^2  {\geq}  \frac{(1 - 2\beta)c_{\mu}^2\Delta_{\min}^2}{4k_{\mu}^2\Vert\xijt\Vert_{\scriptscriptstyle\bM_t}^{-1}} + R^2\log(\frac{\det(M_t)}{\delta_1^2\det(\lambda\mathbf{I})})\right)\\
    \leq & \mathbb{P}\left(\left\| S_t\right\|_{\sub \bM_t^{-1}}^2  {\geq}  2R^2\log\left(\exp\left(\frac{(1 - 2\beta)c_{\mu}^2\Delta_{\min}^2}{8R^2k_{\mu}^2\Vert\xijt\Vert_{\sub \bM_t}^{-1}}\right) \cdot \frac{\det(\bM_t)}{\delta_1^2\det(\lambda\mathbf{I})})\right)\right) \\
    \leq & \delta_1 \cdot \exp^{-1}\left(\frac{(1 - 2\beta)c_{\mu}^2\Delta_{\min}^2}{8R^2k_{\mu}^2\Vert\xijt\Vert_{\sub \bM_t^{-1}}^2}\right) \leq \log(\frac{1}{\delta_1}) \cdot \frac{8R^2k_{\mu}^2\Vert\xijt\Vert_{\bM_t^{-1}}^2}{(1 - 2\beta)c_{\mu}^2\Delta_{\min}^2}
\end{align*}
\end{proof}

\normalsize
\subsection{Proof of Theorem~\ref{theorem}}
With $\delta_1$ and $\delta_2$ defined in the previous lemmas, we have the T-step regret upper bounded as:
\small
\begin{align}
    R_T  = R_{t^\prime} + R_{T - \tp} \leq  R^{\prime} + (1 - \delta_1) (1 - \delta_2)\sum\nolimits_{s = \tp}^{T} r_s 
\label{eqn:regret_all}
 \end{align}
\normalsize
where $R^{\prime} = \tp \cdot L_{\max} + (T - \tp)\left[\delta_2 \cdot L_{\max} + (1 - \delta_2)\cdot \delta_1 L_{\max}\right]$
Under event $E_t^1$ and $E_t^2$, the instantaneous regret at round $s$ is bounded by
\small
\begin{align}
    r_s = \mathbb{E} \big[K(\tau_s, \tau_s^*)\big] = \sum\nolimits_{i=1}^{d_s}\mathbb{E}\left[{(N_i^s + 1)N_i^s}/{2}\right] \leq \mathbb{E}\left[{N_s(N_s + 1)}/{2}\right]
\label{eqn:regret_t}
\end{align}
\normalsize
where $N_i^s$ denotes the number of uncertain rank orders in block $\mathcal{B}_i^s$ at round $s$ and $N_s$ denotes the total number of uncertain rank orders at round $s$. It follows that in the worst case scenario, when $N_s$ uncertain rank orders are all placed into the same block, at most $(N_s^2 + N_s)/2$ mis-ordered pairs will be generated by random shuffling in \model{}. This is due to the fact that based on the block created by \model{}, with $N_s$ uncertain rank orders in one block, this block can have at most $N_s + 1$ documents. 

Therefore, the cumulative regret after $\tp$ can be bounded by:
\small
 \begin{align*}
     \sum\nolimits_{s=\tp}^T r_s \leq & \sum\nolimits_{s=\tp}^T\mathbb{E}\left[{N_s(N_s+1)}/{2}\right] \nonumber \leq  \left(\sum\nolimits_{s=\tp}^T\mathbb{E}[N_s^2] + \mathbb{E}[N_s]\right)/2 \nonumber \\
    \leq &  \left(\mathbb{E}[\sum\nolimits_{s=\tp}^TN_s]^2 +\mathbb{E}[\sum\nolimits_{s=\tp}^TN_s]\right)/2
 \end{align*}
 \normalsize

According to our previous analysis, at round $t \geq \tp$, the number of uncertain rank orders can be estimated by the probability of observing an uncertain rank order, i.e., $\mathbb{P}\big((i, j) \in \mathcal{E}_u^t\big) \leq \log({1}/{\delta_1}) \cdot \frac{8R^2k_{\mu}^2\Vert\xijt\Vert_{M_t^{-1}}^2}{(1 - 2\beta)c_{\mu}^2\Delta_{\min}^2}$. Therefore, the cumulative number of mis-ordered pairs can be bounded by the probability of observing uncertain rank orders in each round, which shrinks with more observations become available over time, 
\small
\begin{align*}
\label{eqn:up}
    \mathbb{E}\left[\sum\nolimits_{s=\tp}^T N_s\right] \leq &  \mathbb{E}\left[\frac{1}{2}\sum\nolimits_{s=\tp}^T \sum\nolimits_{(m, n) \in [L_s]^2} \mathbb{P}\big((m, n) \in \mathcal{E}_u^t\big)\right] \\
    \leq & \mathbb{E}\left[a\sum\nolimits_{s=\tp}^T\sum\nolimits_{(m, n) \in [L_s]^2} \Vert\xmns\Vert_{\bM_s^{-1}}^2\right]
\end{align*}
\normalsize
where $a = \log({1}/{\delta_1})\cdot{4k_{\mu}^2}/({(1 - 2\beta)c_{\mu}^2\Delta_{\min}^2})$

Because $\bM_t$ only contains information of observed document pairs so far, \model{} guarantees the number of mis-ordered pairs among the observed documents in the above inequality is upper bounded. 
To reason about the number of mis-ordered pairs in those unobserved documents (i.e., from $o_t$ to $L_t$ for each query $q_t$), we leverage the constant $p^*$, which is defined as the minimal probability that all documents in a query are examined over time, 
\small
\begin{align*}
    &\mathbb{E}\left[\sum\nolimits_{s=\tp}^T\sum\nolimits_{(m, n) \in [L_s]^2} \Vert\xmns\Vert_{\bM_s^{-1}}^2\right] \\
    =& \mathbb{E}\left[\sum\nolimits_{s=\tp}^T\sum\nolimits_{(m, n) \in [L_s]^2} \Vert\xmns\Vert_{\bM_s^{-1}}^2\times \mathbb{E}\left[\frac{1}{p_{s}}\mathbf{1}\{o_s = L_s\}\right]\right] \\
    \leq& 
    {p^*}^{-1}\mathbb{E}\left[\sum\nolimits_{s=\tp}^T\sum\nolimits_{(m, n) \in [L_s]^2} \Vert\xmns\Vert_{\bM_s^{-1}}^2 \mathbf{1}\{o_s = L_s\}\right] 
\end{align*}
\normalsize

Besides, in \model{}, we only use the independent pairs, $\mathcal{G}_t^{ind}$ to update the model and the corresponding $M_t$ matrix. Therefore, to bound the regret, we rewrite the above equation as:

\small
\begin{align*}
    &\mathbb{E}\left[\sum\nolimits_{s=\tp}^T\sum\nolimits_{(m, n) \in [L_s]^2} \Vert\xmns\Vert_{\bM_s^{-1}}^2\right]  \\
    =& \mathbb{E}\left[\sum_{s=\tp}^T\sum_{(m, n) \in \mathcal{G}_t^{ind}} \left (\Vert\xmns\Vert_{\bM_s^{-1}}^2 + \sum_{k\in [L_t] \setminus \{m, n\} } \Vert\bx_{mk}^s\Vert_{\bM_s^{-1}}^2 + \Vert\bx_{nk}^s\Vert_{\bM_s^{-1}}^2\right)\right] \\
    =& \mathbb{E}\left[\sum_{s=\tp}^T\sum_{(m, n) \in \mathcal{G}_t^{ind}} \left ((L_t - 1)\Vert\xmns\Vert_{\bM_s^{-1}}^2 + \sum_{k\in [L_t] \setminus \{m, n\} } 2{\bx_{mk}^s}^\top\bM_s^{-1}\bx_{nk}^s\right)\right]
\end{align*}
\normalsize

Based on Lemma 10 and Lemma 11 in \cite{abbasi2011improved}, the expected number of the first term can be upper bounded by,
\small
\begin{align*}
    \sum\nolimits_{s=\tp}^T\sum\nolimits_{(m, n) \in \mathcal{G}_t^{ind}} (L_t - 1)\Vert\xmns\Vert_{\bM_s^{-1}}^2 \leq 2dL_{max}\log(1 + \frac{o_{\max}Tu^2}{2d\lambda})
\end{align*}
\normalsize
And the second term can be bounded as:
\small
\begin{align*}
    &\sum\nolimits_{s=\tp}^T\sum\nolimits_{(m, n) \in \mathcal{G}_t^{ind}}\sum\nolimits_{k\in [L_t] \setminus \{m, n\} } 2{\bx_{mk}^s}^\top\bM_s^{-1}\bx_{nk}^s \\
    \leq & \sum\nolimits_{s=\tp}^T {(L_{max}^2 - 2L_{max})u^2 }/{\lambda_{\min}(\bM_s)} = w.
\end{align*}
\normalsize

Therefore, combining the conclusion of $\mathbb{E}[\sum_{s = \tp}^T N_s]$, and Eq~\eqref{eqn:regret_all}, Eq~\eqref{eqn:regret_t}, the regret could be upper bounded by:
 \small
 \begin{align*}
     R_T 
     \leq& R^{\prime} + (1 - \delta_1)(1 - \delta_2) 
     \frac{1}{p^*{^2}}\left( 2adL_{\max}\log(1 + \frac{o_{\max}Tu^2}{2d\lambda}) + aw\right)^2
 \end{align*}
 \normalsize
 
 
where $R^{\prime} = \tp L_{\max} + (T - t')(\delta_2L_{\max} - (1- \delta_2)\delta_1 L_{\max})$, with $\tp$ defined in Lemma \ref{lemma:uncertain}, and $L_{\max}$ representing the maximum number of documents associated to the same query over time. 
By choosing $\delta_1 = \delta_2 =1/T$, the theorem shows that the expected regret is at most $R_T \leq O(d\log^4(T))$.


%% file: www_2021.bbl

\begin{thebibliography}{38}


\ifx \showCODEN    \undefined \def \showCODEN     #1{\unskip}     \fi
\ifx \showDOI      \undefined \def \showDOI       #1{#1}\fi
\ifx \showISBNx    \undefined \def \showISBNx     #1{\unskip}     \fi
\ifx \showISBNxiii \undefined \def \showISBNxiii  #1{\unskip}     \fi
\ifx \showISSN     \undefined \def \showISSN      #1{\unskip}     \fi
\ifx \showLCCN     \undefined \def \showLCCN      #1{\unskip}     \fi
\ifx \shownote     \undefined \def \shownote      #1{#1}          \fi
\ifx \showarticletitle \undefined \def \showarticletitle #1{#1}   \fi
\ifx \showURL      \undefined \def \showURL       {\relax}        \fi
\providecommand\bibfield[2]{#2}
\providecommand\bibinfo[2]{#2}
\providecommand\natexlab[1]{#1}
\providecommand\showeprint[2][]{arXiv:#2}

\bibitem[\protect\citeauthoryear{Abbasi-Yadkori, P{\'a}l, and
  Szepesv{\'a}ri}{Abbasi-Yadkori et~al\mbox{.}}{2011}]%
        {abbasi2011improved}
\bibfield{author}{\bibinfo{person}{Yasin Abbasi-Yadkori},
  \bibinfo{person}{D{\'a}vid P{\'a}l}, {and} \bibinfo{person}{Csaba
  Szepesv{\'a}ri}.} \bibinfo{year}{2011}\natexlab{}.
\newblock \showarticletitle{Improved algorithms for linear stochastic bandits}.
  In \bibinfo{booktitle}{\emph{NIPS}}. \bibinfo{pages}{2312--2320}.
\newblock


\bibitem[\protect\citeauthoryear{Agichtein, Brill, and Dumais}{Agichtein
  et~al\mbox{.}}{2006}]%
        {agichtein2006improving}
\bibfield{author}{\bibinfo{person}{Eugene Agichtein}, \bibinfo{person}{Eric
  Brill}, {and} \bibinfo{person}{Susan Dumais}.}
  \bibinfo{year}{2006}\natexlab{}.
\newblock \showarticletitle{Improving web search ranking by incorporating user
  behavior information}. In \bibinfo{booktitle}{\emph{Proceedings of the 29th
  ACM SIGIR}}. ACM, \bibinfo{pages}{19--26}.
\newblock


\bibitem[\protect\citeauthoryear{Burges}{Burges}{2010}]%
        {burges2010ranknet}
\bibfield{author}{\bibinfo{person}{Christopher~JC Burges}.}
  \bibinfo{year}{2010}\natexlab{}.
\newblock \showarticletitle{From ranknet to lambdarank to lambdamart: An
  overview}.
\newblock \bibinfo{journal}{\emph{Learning}} \bibinfo{volume}{11},
  \bibinfo{number}{23-581} (\bibinfo{year}{2010}), \bibinfo{pages}{81}.
\newblock


\bibitem[\protect\citeauthoryear{Carvalho, Elsas, Cohen, and
  Carbonell}{Carvalho et~al\mbox{.}}{2008}]%
        {carvalho2008suppressing}
\bibfield{author}{\bibinfo{person}{Vitor~R Carvalho},
  \bibinfo{person}{Jonathan~L Elsas}, \bibinfo{person}{William~W Cohen}, {and}
  \bibinfo{person}{Jaime~G Carbonell}.} \bibinfo{year}{2008}\natexlab{}.
\newblock \showarticletitle{Suppressing outliers in pairwise preference
  ranking}. In \bibinfo{booktitle}{\emph{Proceedings of the 17th ACM CIKM}}.
  \bibinfo{pages}{1487--1488}.
\newblock


\bibitem[\protect\citeauthoryear{Chapelle and Chang}{Chapelle and
  Chang}{2011}]%
        {chapelle2011yahoo}
\bibfield{author}{\bibinfo{person}{Olivier Chapelle} {and} \bibinfo{person}{Yi
  Chang}.} \bibinfo{year}{2011}\natexlab{}.
\newblock \showarticletitle{Yahoo! learning to rank challenge overview}. In
  \bibinfo{booktitle}{\emph{Proceedings of the Learning to Rank Challenge}}.
  \bibinfo{pages}{1--24}.
\newblock


\bibitem[\protect\citeauthoryear{Chapelle, Joachims, Radlinski, and
  Yue}{Chapelle et~al\mbox{.}}{2012}]%
        {chapelle2012large}
\bibfield{author}{\bibinfo{person}{Olivier Chapelle}, \bibinfo{person}{Thorsten
  Joachims}, \bibinfo{person}{Filip Radlinski}, {and} \bibinfo{person}{Yisong
  Yue}.} \bibinfo{year}{2012}\natexlab{}.
\newblock \showarticletitle{Large-scale validation and analysis of interleaved
  search evaluation}.
\newblock \bibinfo{journal}{\emph{ACM TOIS}} \bibinfo{volume}{30},
  \bibinfo{number}{1} (\bibinfo{year}{2012}), \bibinfo{pages}{6}.
\newblock


\bibitem[\protect\citeauthoryear{Filippi, Cappe, Garivier, and
  Szepesv{\'a}ri}{Filippi et~al\mbox{.}}{2010}]%
        {filippi2010parametric}
\bibfield{author}{\bibinfo{person}{Sarah Filippi}, \bibinfo{person}{Olivier
  Cappe}, \bibinfo{person}{Aur{\'e}lien Garivier}, {and} \bibinfo{person}{Csaba
  Szepesv{\'a}ri}.} \bibinfo{year}{2010}\natexlab{}.
\newblock \showarticletitle{Parametric bandits: The generalized linear case}.
  In \bibinfo{booktitle}{\emph{NIPS}}. \bibinfo{pages}{586--594}.
\newblock


\bibitem[\protect\citeauthoryear{Guo, Liu, Kannan, Minka, Taylor, Wang, and
  Faloutsos}{Guo et~al\mbox{.}}{2009b}]%
        {guo2009click}
\bibfield{author}{\bibinfo{person}{Fan Guo}, \bibinfo{person}{Chao Liu},
  \bibinfo{person}{Anitha Kannan}, \bibinfo{person}{Tom Minka},
  \bibinfo{person}{Michael Taylor}, \bibinfo{person}{Yi-Min Wang}, {and}
  \bibinfo{person}{Christos Faloutsos}.} \bibinfo{year}{2009}\natexlab{b}.
\newblock \showarticletitle{Click chain model in web search}. In
  \bibinfo{booktitle}{\emph{Proceedings of the 18th WWW}}.
  \bibinfo{pages}{11--20}.
\newblock


\bibitem[\protect\citeauthoryear{Guo, Liu, and Wang}{Guo
  et~al\mbox{.}}{2009a}]%
        {guo2009efficient}
\bibfield{author}{\bibinfo{person}{Fan Guo}, \bibinfo{person}{Chao Liu}, {and}
  \bibinfo{person}{Yi~Min Wang}.} \bibinfo{year}{2009}\natexlab{a}.
\newblock \showarticletitle{Efficient multiple-click models in web search}. In
  \bibinfo{booktitle}{\emph{Proceedings of the 2nd WSDM}}.
  \bibinfo{pages}{124--131}.
\newblock


\bibitem[\protect\citeauthoryear{Herbrich, Graepel, and Obermayer}{Herbrich
  et~al\mbox{.}}{1999}]%
        {herbrich1999support}
\bibfield{author}{\bibinfo{person}{Ralf Herbrich}, \bibinfo{person}{Thore
  Graepel}, {and} \bibinfo{person}{Klaus Obermayer}.}
  \bibinfo{year}{1999}\natexlab{}.
\newblock \showarticletitle{Support vector learning for ordinal regression}.
\newblock  (\bibinfo{year}{1999}).
\newblock


\bibitem[\protect\citeauthoryear{Hofmann, Whiteson, and de~Rijke}{Hofmann
  et~al\mbox{.}}{2012}]%
        {hofmann2012estimating}
\bibfield{author}{\bibinfo{person}{Katja Hofmann}, \bibinfo{person}{Shimon
  Whiteson}, {and} \bibinfo{person}{Maarten de Rijke}.}
  \bibinfo{year}{2012}\natexlab{}.
\newblock \showarticletitle{Estimating interleaved comparison outcomes from
  historical click data}. In \bibinfo{booktitle}{\emph{Proceedings of the 21st
  CIKM}}. \bibinfo{pages}{1779--1783}.
\newblock


\bibitem[\protect\citeauthoryear{Hofmann, Whiteson, and de~Rijke}{Hofmann
  et~al\mbox{.}}{2013}]%
        {hofmann2013balancing}
\bibfield{author}{\bibinfo{person}{Katja Hofmann}, \bibinfo{person}{Shimon
  Whiteson}, {and} \bibinfo{person}{Maarten de Rijke}.}
  \bibinfo{year}{2013}\natexlab{}.
\newblock \showarticletitle{Balancing exploration and exploitation in listwise
  and pairwise online learning to rank for information retrieval}.
\newblock \bibinfo{journal}{\emph{Information Retrieval}} \bibinfo{volume}{16},
  \bibinfo{number}{1} (\bibinfo{year}{2013}), \bibinfo{pages}{63--90}.
\newblock


\bibitem[\protect\citeauthoryear{Joachims}{Joachims}{2002}]%
        {joachims2002optimizing}
\bibfield{author}{\bibinfo{person}{Thorsten Joachims}.}
  \bibinfo{year}{2002}\natexlab{}.
\newblock \showarticletitle{Optimizing search engines using clickthrough data}.
  In \bibinfo{booktitle}{\emph{Proceedings of the 8th ACM KDD}}.
  \bibinfo{pages}{133--142}.
\newblock


\bibitem[\protect\citeauthoryear{Joachims, Granka, Pan, Hembrooke, and
  Gay}{Joachims et~al\mbox{.}}{2005}]%
        {joachims2005accurately}
\bibfield{author}{\bibinfo{person}{Thorsten Joachims}, \bibinfo{person}{Laura
  Granka}, \bibinfo{person}{Bing Pan}, \bibinfo{person}{Helene Hembrooke},
  {and} \bibinfo{person}{Geri Gay}.} \bibinfo{year}{2005}\natexlab{}.
\newblock \showarticletitle{Accurately interpreting clickthrough data as
  implicit feedback}. In \bibinfo{booktitle}{\emph{Proceedings of the 28th ACM
  SIGIR}}. ACM, \bibinfo{pages}{154--161}.
\newblock


\bibitem[\protect\citeauthoryear{Joachims, Granka, Pan, Hembrooke, Radlinski,
  and Gay}{Joachims et~al\mbox{.}}{2007}]%
        {joachims2007evaluating}
\bibfield{author}{\bibinfo{person}{Thorsten Joachims}, \bibinfo{person}{Laura
  Granka}, \bibinfo{person}{Bing Pan}, \bibinfo{person}{Helene Hembrooke},
  \bibinfo{person}{Filip Radlinski}, {and} \bibinfo{person}{Geri Gay}.}
  \bibinfo{year}{2007}\natexlab{}.
\newblock \showarticletitle{Evaluating the accuracy of implicit feedback from
  clicks and query reformulations in web search}.
\newblock \bibinfo{journal}{\emph{ACM TOIS}} \bibinfo{volume}{25},
  \bibinfo{number}{2} (\bibinfo{year}{2007}), \bibinfo{pages}{7}.
\newblock


\bibitem[\protect\citeauthoryear{Joachims, Swaminathan, and Schnabel}{Joachims
  et~al\mbox{.}}{2017}]%
        {joachims2017ips}
\bibfield{author}{\bibinfo{person}{Thorsten Joachims}, \bibinfo{person}{Adith
  Swaminathan}, {and} \bibinfo{person}{Tobias Schnabel}.}
  \bibinfo{year}{2017}\natexlab{}.
\newblock \showarticletitle{Unbiased Learning-to-Rank with Biased Feedback}. In
  \bibinfo{booktitle}{\emph{Proceedings of the 10th ACM WSDM}}.
  \bibinfo{publisher}{ACM}, \bibinfo{pages}{781--789}.
\newblock


\bibitem[\protect\citeauthoryear{Katariya, Kveton, Szepesvari, and
  Wen}{Katariya et~al\mbox{.}}{2016}]%
        {katariya2016dcm}
\bibfield{author}{\bibinfo{person}{Sumeet Katariya}, \bibinfo{person}{Branislav
  Kveton}, \bibinfo{person}{Csaba Szepesvari}, {and} \bibinfo{person}{Zheng
  Wen}.} \bibinfo{year}{2016}\natexlab{}.
\newblock \showarticletitle{DCM bandits: Learning to rank with multiple
  clicks}. In \bibinfo{booktitle}{\emph{ICML}}. \bibinfo{pages}{1215--1224}.
\newblock


\bibitem[\protect\citeauthoryear{Kveton, Li, Lattimore, Markov, de~Rijke,
  Szepesvari, and Zoghi}{Kveton et~al\mbox{.}}{2018}]%
        {kveton2018bubblerank}
\bibfield{author}{\bibinfo{person}{Branislav Kveton}, \bibinfo{person}{Chang
  Li}, \bibinfo{person}{Tor Lattimore}, \bibinfo{person}{Ilya Markov},
  \bibinfo{person}{Maarten de Rijke}, \bibinfo{person}{Csaba Szepesvari}, {and}
  \bibinfo{person}{Masrour Zoghi}.} \bibinfo{year}{2018}\natexlab{}.
\newblock \showarticletitle{Bubblerank: Safe online learning to rerank}.
\newblock \bibinfo{journal}{\emph{arXiv preprint arXiv:1806.05819}}
  (\bibinfo{year}{2018}).
\newblock


\bibitem[\protect\citeauthoryear{Kveton, Szepesvari, Wen, and Ashkan}{Kveton
  et~al\mbox{.}}{2015a}]%
        {kveton2015cascading}
\bibfield{author}{\bibinfo{person}{Branislav Kveton}, \bibinfo{person}{Csaba
  Szepesvari}, \bibinfo{person}{Zheng Wen}, {and} \bibinfo{person}{Azin
  Ashkan}.} \bibinfo{year}{2015}\natexlab{a}.
\newblock \showarticletitle{Cascading bandits: Learning to rank in the cascade
  model}. In \bibinfo{booktitle}{\emph{ICML}}. \bibinfo{pages}{767--776}.
\newblock


\bibitem[\protect\citeauthoryear{Kveton, Wen, Ashkan, and Szepesvari}{Kveton
  et~al\mbox{.}}{2015b}]%
        {kveton2015combinatorial}
\bibfield{author}{\bibinfo{person}{Branislav Kveton}, \bibinfo{person}{Zheng
  Wen}, \bibinfo{person}{Azin Ashkan}, {and} \bibinfo{person}{Csaba
  Szepesvari}.} \bibinfo{year}{2015}\natexlab{b}.
\newblock \showarticletitle{Combinatorial cascading bandits}. In
  \bibinfo{booktitle}{\emph{NIPS}}. \bibinfo{pages}{1450--1458}.
\newblock


\bibitem[\protect\citeauthoryear{Kveton, Wen, Ashkan, and Szepesvari}{Kveton
  et~al\mbox{.}}{2015c}]%
        {kveton2015tight}
\bibfield{author}{\bibinfo{person}{Branislav Kveton}, \bibinfo{person}{Zheng
  Wen}, \bibinfo{person}{Azin Ashkan}, {and} \bibinfo{person}{Csaba
  Szepesvari}.} \bibinfo{year}{2015}\natexlab{c}.
\newblock \showarticletitle{Tight regret bounds for stochastic combinatorial
  semi-bandits}. In \bibinfo{booktitle}{\emph{Artificial Intelligence and
  Statistics}}. \bibinfo{pages}{535--543}.
\newblock


\bibitem[\protect\citeauthoryear{Lattimore, Kveton, Li, and
  Szepesvari}{Lattimore et~al\mbox{.}}{2018}]%
        {lattimore2018toprank}
\bibfield{author}{\bibinfo{person}{Tor Lattimore}, \bibinfo{person}{Branislav
  Kveton}, \bibinfo{person}{Shuai Li}, {and} \bibinfo{person}{Csaba
  Szepesvari}.} \bibinfo{year}{2018}\natexlab{}.
\newblock \showarticletitle{Toprank: A practical algorithm for online
  stochastic ranking}. In \bibinfo{booktitle}{\emph{NIPS}}.
  \bibinfo{pages}{3945--3954}.
\newblock


\bibitem[\protect\citeauthoryear{Li, Lattimore, and Szepesv{\'a}ri}{Li
  et~al\mbox{.}}{2018}]%
        {li2018online}
\bibfield{author}{\bibinfo{person}{Shuai Li}, \bibinfo{person}{Tor Lattimore},
  {and} \bibinfo{person}{Csaba Szepesv{\'a}ri}.}
  \bibinfo{year}{2018}\natexlab{}.
\newblock \showarticletitle{Online learning to rank with features}.
\newblock \bibinfo{journal}{\emph{arXiv preprint arXiv:1810.02567}}
  (\bibinfo{year}{2018}).
\newblock


\bibitem[\protect\citeauthoryear{Li, Wang, Zhang, and Chen}{Li
  et~al\mbox{.}}{2016}]%
        {li2016contextual}
\bibfield{author}{\bibinfo{person}{Shuai Li}, \bibinfo{person}{Baoxiang Wang},
  \bibinfo{person}{Shengyu Zhang}, {and} \bibinfo{person}{Wei Chen}.}
  \bibinfo{year}{2016}\natexlab{}.
\newblock \showarticletitle{Contextual Combinatorial Cascading Bandits.}. In
  \bibinfo{booktitle}{\emph{ICML}}, Vol.~\bibinfo{volume}{16}.
  \bibinfo{pages}{1245--1253}.
\newblock


\bibitem[\protect\citeauthoryear{Oosterhuis and de~Rijke}{Oosterhuis and
  de~Rijke}{2017}]%
        {oosterhuis2017balancing}
\bibfield{author}{\bibinfo{person}{Harrie Oosterhuis} {and}
  \bibinfo{person}{Maarten de Rijke}.} \bibinfo{year}{2017}\natexlab{}.
\newblock \showarticletitle{Balancing speed and quality in online learning to
  rank for information retrieval}. In \bibinfo{booktitle}{\emph{Proceedings of
  the 26th 2017 ACM CIKM}}. \bibinfo{pages}{277--286}.
\newblock


\bibitem[\protect\citeauthoryear{Oosterhuis and de~Rijke}{Oosterhuis and
  de~Rijke}{2018}]%
        {oosterhuis2018differentiable}
\bibfield{author}{\bibinfo{person}{Harrie Oosterhuis} {and}
  \bibinfo{person}{Maarten de Rijke}.} \bibinfo{year}{2018}\natexlab{}.
\newblock \showarticletitle{Differentiable unbiased online learning to rank}.
  In \bibinfo{booktitle}{\emph{Proceedings of the 27th ACM CIKM}}.
  \bibinfo{pages}{1293--1302}.
\newblock


\bibitem[\protect\citeauthoryear{Qin and Liu}{Qin and Liu}{2013}]%
        {qin2013introducing}
\bibfield{author}{\bibinfo{person}{Tao Qin} {and} \bibinfo{person}{Tie-Yan
  Liu}.} \bibinfo{year}{2013}\natexlab{}.
\newblock \bibinfo{title}{Introducing LETOR 4.0 Datasets}.
\newblock
\newblock
\showeprint[arxiv]{1306.2597}~[cs.IR]


\bibitem[\protect\citeauthoryear{Radlinski, Kleinberg, and Joachims}{Radlinski
  et~al\mbox{.}}{2008}]%
        {radlinski2008learning}
\bibfield{author}{\bibinfo{person}{Filip Radlinski}, \bibinfo{person}{Robert
  Kleinberg}, {and} \bibinfo{person}{Thorsten Joachims}.}
  \bibinfo{year}{2008}\natexlab{}.
\newblock \showarticletitle{Learning diverse rankings with multi-armed
  bandits}. In \bibinfo{booktitle}{\emph{ICML}}. \bibinfo{pages}{784--791}.
\newblock


\bibitem[\protect\citeauthoryear{Schuth, Oosterhuis, Whiteson, and
  de~Rijke}{Schuth et~al\mbox{.}}{2016}]%
        {schuth2016multileave}
\bibfield{author}{\bibinfo{person}{Anne Schuth}, \bibinfo{person}{Harrie
  Oosterhuis}, \bibinfo{person}{Shimon Whiteson}, {and}
  \bibinfo{person}{Maarten de Rijke}.} \bibinfo{year}{2016}\natexlab{}.
\newblock \showarticletitle{Multileave gradient descent for fast online
  learning to rank}. In \bibinfo{booktitle}{\emph{Proceedings of the 9th ACM
  WSDM}}. \bibinfo{pages}{457--466}.
\newblock


\bibitem[\protect\citeauthoryear{Schuth, Sietsma, Whiteson, Lefortier, and
  de~Rijke}{Schuth et~al\mbox{.}}{2014}]%
        {schuth2014multileaved}
\bibfield{author}{\bibinfo{person}{Anne Schuth}, \bibinfo{person}{Floor
  Sietsma}, \bibinfo{person}{Shimon Whiteson}, \bibinfo{person}{Damien
  Lefortier}, {and} \bibinfo{person}{Maarten de Rijke}.}
  \bibinfo{year}{2014}\natexlab{}.
\newblock \showarticletitle{Multileaved comparisons for fast online
  evaluation}. In \bibinfo{booktitle}{\emph{Proceedings of the 23rd ACM CIKM}}.
  ACM, \bibinfo{pages}{71--80}.
\newblock


\bibitem[\protect\citeauthoryear{Vershynin}{Vershynin}{2010}]%
        {vershynin2010introduction}
\bibfield{author}{\bibinfo{person}{Roman Vershynin}.}
  \bibinfo{year}{2010}\natexlab{}.
\newblock \showarticletitle{Introduction to the non-asymptotic analysis of
  random matrices}.
\newblock \bibinfo{journal}{\emph{arXiv preprint arXiv:1011.3027}}
  (\bibinfo{year}{2010}).
\newblock


\bibitem[\protect\citeauthoryear{Wang, Kim, McCord-Snook, Wu, and Wang}{Wang
  et~al\mbox{.}}{2019}]%
        {wang2019variance}
\bibfield{author}{\bibinfo{person}{Huazheng Wang}, \bibinfo{person}{Sonwoo
  Kim}, \bibinfo{person}{Eric McCord-Snook}, \bibinfo{person}{Qingyun Wu},
  {and} \bibinfo{person}{Hongning Wang}.} \bibinfo{year}{2019}\natexlab{}.
\newblock \showarticletitle{Variance Reduction in Gradient Exploration for
  Online Learning to Rank}. In \bibinfo{booktitle}{\emph{SIGIR 2019}}.
  \bibinfo{pages}{835--844}.
\newblock


\bibitem[\protect\citeauthoryear{Wang, Langley, Kim, McCord-Snook, and
  Wang}{Wang et~al\mbox{.}}{2018a}]%
        {wang2018efficient}
\bibfield{author}{\bibinfo{person}{Huazheng Wang}, \bibinfo{person}{Ramsey
  Langley}, \bibinfo{person}{Sonwoo Kim}, \bibinfo{person}{Eric McCord-Snook},
  {and} \bibinfo{person}{Hongning Wang}.} \bibinfo{year}{2018}\natexlab{a}.
\newblock \showarticletitle{Efficient exploration of gradient space for online
  learning to rank}. In \bibinfo{booktitle}{\emph{SIGIR 2018}}.
  \bibinfo{pages}{145--154}.
\newblock


\bibitem[\protect\citeauthoryear{Wang, Li, Golbandi, Bendersky, and
  Najork}{Wang et~al\mbox{.}}{2018b}]%
        {Wang2018Lambdaloss}
\bibfield{author}{\bibinfo{person}{Xuanhui Wang}, \bibinfo{person}{Cheng Li},
  \bibinfo{person}{Nadav Golbandi}, \bibinfo{person}{Michael Bendersky}, {and}
  \bibinfo{person}{Marc Najork}.} \bibinfo{year}{2018}\natexlab{b}.
\newblock \showarticletitle{The LambdaLoss Framework for Ranking Metric
  Optimization}. In \bibinfo{booktitle}{\emph{CIKM '18}}.
  \bibinfo{publisher}{ACM}, \bibinfo{pages}{1313--1322}.
\newblock


\bibitem[\protect\citeauthoryear{Yue and Joachims}{Yue and Joachims}{2009}]%
        {yue2009interactively}
\bibfield{author}{\bibinfo{person}{Yisong Yue} {and} \bibinfo{person}{Thorsten
  Joachims}.} \bibinfo{year}{2009}\natexlab{}.
\newblock \showarticletitle{Interactively optimizing information retrieval
  systems as a dueling bandits problem}. In \bibinfo{booktitle}{\emph{ICML}}.
  \bibinfo{pages}{1201--1208}.
\newblock


\bibitem[\protect\citeauthoryear{Zhao and King}{Zhao and King}{2016}]%
        {zhao2016constructing}
\bibfield{author}{\bibinfo{person}{Tong Zhao} {and} \bibinfo{person}{Irwin
  King}.} \bibinfo{year}{2016}\natexlab{}.
\newblock \showarticletitle{Constructing reliable gradient exploration for
  online learning to rank}. In \bibinfo{booktitle}{\emph{Proceedings of the
  25th ACM CIKM}}. \bibinfo{pages}{1643--1652}.
\newblock


\bibitem[\protect\citeauthoryear{Zhou, Li, and Gu}{Zhou et~al\mbox{.}}{2019}]%
        {zhou2019neural}
\bibfield{author}{\bibinfo{person}{Dongruo Zhou}, \bibinfo{person}{Lihong Li},
  {and} \bibinfo{person}{Quanquan Gu}.} \bibinfo{year}{2019}\natexlab{}.
\newblock \showarticletitle{Neural Contextual Bandits with UCB-based
  Exploration}.
\newblock \bibinfo{journal}{\emph{arXiv preprint arXiv:1911.04462}}
  (\bibinfo{year}{2019}).
\newblock


\bibitem[\protect\citeauthoryear{Zoghi, Tunys, Ghavamzadeh, Kveton, Szepesvari,
  and Wen}{Zoghi et~al\mbox{.}}{2017}]%
        {zoghi2017online}
\bibfield{author}{\bibinfo{person}{Masrour Zoghi}, \bibinfo{person}{Tomas
  Tunys}, \bibinfo{person}{Mohammad Ghavamzadeh}, \bibinfo{person}{Branislav
  Kveton}, \bibinfo{person}{Csaba Szepesvari}, {and} \bibinfo{person}{Zheng
  Wen}.} \bibinfo{year}{2017}\natexlab{}.
\newblock \showarticletitle{Online learning to rank in stochastic click
  models}. In \bibinfo{booktitle}{\emph{ICML 2017}}.
  \bibinfo{pages}{4199--4208}.
\newblock


\end{thebibliography}
